\documentclass[AMA,Times1COL]{WileyNJDv5}

\usepackage{multirow}
\usepackage{colortbl}
\usepackage{xcolor}
\usepackage{orcidlink}
\usepackage{hyperref}
\usepackage{orcidlink}
\usepackage{diagbox}

\usepackage[thicklines]{cancel}

\articletype{Research Article - Methodology}%

\received{00} 
\revised{00} 
\accepted{00}
\journal{Journal}
\volume{00}
\copyyear{2023}
\startpage{1}

\begin{document}

\title{Better Knowledge Enhancement for Privacy-Preserving Cross-Project Defect Prediction}

\author[1]{Yuying Wang}
% \author[2]{Yichen Li}
\author{Yichen Li~\orcidlink{0009-0009-8630-2504}$^\textbf{2}$}
\author[2]{Haozhao Wang}
\author[1]{Lei Zhao}
\author[1]{Xiaofang Zhang~\orcidlink{0000-0002-8667-0456}}

\authormark{Wang, Yuying, et al.}
\titlemark{Better Knowledge Enhancement for Privacy-Preserving Cross-Project Defect Prediction}

\address[1]{\orgdiv{School of Computer Science and Technology}, \orgname{Soochow University}, \orgaddress{\state{Suzhou}, \country{China}}}

\address[2]{\orgdiv{School of Computer Science and Technology}, \orgname{Huazhong University of Science and Technology}, \orgaddress{\state{Wuhan}, \country{China}}}

\corres{Xiaofang Zhang, School of Computer Science and Technology, Soochow University, Suzhou,215000, China. \email{xfzhang@suda.edu.cn}
\\Yichen Li, School of Computer Science and Technology, Huazhong University of Science and Technology, Wuhan,430000, China. \email{ycli0204@hust.edu.cn}
}

% \presentaddress{This is sample for present address text this is sample for present address text.}

%\fundingInfo{Text}
%\JELinfo{ejlje}

\newcommand{\minew}[1]{{\color{blue}{#1}}}
\newcommand{\miold}[1]{{\color{red}\sout{#1}}}
\renewcommand{\CancelColor}{\color{red}}
\abstract[Abstract]{Cross-Project Defect Prediction (CPDP) poses a non-trivial challenge to construct a reliable defect predictor by leveraging data from other projects, particularly when data owners are concerned about data privacy. In recent years, Federated Learning (FL) has become an emerging paradigm to guarantee privacy information by collaborative training a global model among multiple parties without sharing raw data. 
While the direct application of FL to the CPDP task offers a promising solution to address privacy concerns, the data heterogeneity arising from proprietary projects across different companies or organizations will bring troubles for model training. 
In this paper, we study the privacy-preserving cross-project defect prediction with data heterogeneity under the federated learning framework. 
% Open-source project data are effective for centralized CPDP with data heterogeneity to enhance model performance.
% However, we fail when trying to intuitively consolidate the global model by additionally training on open-source project data at the server, where the unstable training process with uneven open-source project data leads to notable performance degradation. 
To address this problem, we propose a novel knowledge enhancement approach named \textbf{FedDP} with two simple but effective solutions: 1. Local Heterogeneity Awareness and 2. Global Knowledge Distillation. Specifically, we employ open-source project data as the distillation dataset and optimize the global model with the heterogeneity-aware local model ensemble via knowledge distillation. 
Experimental results on 19 projects from two datasets demonstrate that our method significantly outperforms baselines.}


% \jnlcitation{\cname{%
% \author{Taylor M.},
% \author{Lauritzen P},
% \author{Erath C}, and
% \author{Mittal R}}.
% \ctitle{On simplifying ‘incremental remap’-based transport schemes.} \cjournal{\it J Comput Phys.} \cvol{2021;00(00):1--18}.}

\maketitle

% \renewcommand\thefootnote{}
% % : 这将脚注的计数器格式设置为空，意味着接下来的脚注将没有数字或符号。这通常用于特殊的脚注，如标题页的版权声明或作者注释。
% \footnotetext{\textbf{Abbreviations:} ANA, anti-nuclear antibodies; APC, antigen-presenting cells; IRF, interferon regulatory factor.}
% % 使用没有标记的脚注来添加一组缩写说明。由于\thefootnote已经被设置为空，这个脚注将没有可见的标记。

\renewcommand\thefootnote{\fnsymbol{footnote}}
% 这将脚注的计数器格式设置为使用符号（例如，星号、圆圈等），而不是数字。\fnsymbol{footnote}是一个LaTeX命令，它根据当前的脚注计数器值生成一个符号。
\setcounter{footnote}{1}
% 这将脚注计数器重置为1。由于之前的脚注没有使用计数器（因为它没有可见的标记），所以这个命令确保下一个脚注从符号1开始（通常是星号）。

\section{INTRODUCTION}\label{intro}
In the field of software engineering, Software Defect Prediction (SDP) is considered a crucial task. The primary goal of SDP is to identify potential defects at the early stages of software development, which serves to enhance the quality of the software and reduce maintenance costs \cite{SDP1,SDP2,SDP3}. Currently, SDP mainly encompasses two typical scenarios: Within-Project Defect Prediction (WPDP) and Cross-Project Defect Prediction (CPDP). Most existing works focus on WPDP and assume that target projects possess ample historical data for training, enabling the identification of potentially defective modules within the new versions of the same project's data \cite{WPDP1,WPDP2,WPDP3}.

In practice, the collection of defect data has historically been deemed a protracted and intricate task\cite{cpdp_nec1}. As a result, not all projects have sufficient historical data for training, especially for newly developed projects\cite{CPDP1}. Even for the same project, with evolving technologies and changes in business environments, previously gathered data may no longer be fully applicable to current data \cite{cpdp_nec3}. Models trained solely on outdated within-project data may struggle to adapt to these changes, resulting in a decline in prediction performance. Against this backdrop, CPDP emerges as a viable option for companies. CPDP leverages additional data from other projects to train a defect predictor for target projects\cite{CPDP2, CPDP3}, significantly reducing the workload associated with data re-collection and reorganization\cite{cpdp_nec2}. And multi-source cross-project training leverages data from multiple projects simultaneously, enhancing data diversity. This diversity enables the model to grasp a broader range of defect patterns and features, surpassing the limitations of the single project.
% Especially in newly developed projects, these projects often lack sufficient historical data to support model training in real-world scenarios. 

% where historical data may be limited or non-existent. 
% CPDP has been introduced to address this challenge by utilizing additional data from other projects to train a defect predictor for target projects\cite{CPDP1, CPDP2, CPDP3}.

% Previous research has commonly relied on open-source projects (e.g., Apache) or repositories (e.g., GitHub) for CPDP studies. Nevertheless, models constructed solely from open-source data may struggle to be effectively applied to proprietary projects\cite{xiaxin}. Software development is influenced by many factors, including geography, organizational structure, and project characteristics. For example, Fendler et al. \cite{datadis1} observed the limitations of applying Western software engineering practices in African contexts. Similarly, Unterkalmsteiner et al. \cite{datadis2} found substantial differences in software development between startups and mature large-scale companies. Open-source project data may not adequately capture these nuances\cite{Shriram}. 

However, industrial data frequently contains a substantial amount of sensitive information, and the disclosure of such information may jeopardize corporate business interests. Proprietary data owners' concerns about privacy often pose obstacles to data sharing, thereby hindering the practical implementation of CPDP in real-world scenarios. For example, AT\&T, a global telecom giant, hesitated to publicize data due to fears of exposing sensitive business details \cite{Weyuker}. Similarly, revealing metrics like code complexity and the lines of code can give competitors insights into development efficiency, threatening a company's reputation and pricing power \cite{Li_loc,PetersLACE2}. To overcome this problem, some researchers have explored ways to share project data in CPDP without leaking sensitive information. Conventionally, LACE2 \cite{PetersLACE2} implements a multi-party data-sharing mechanism that reduces the total amount of data needed to be shared. But LACE2 still needs to share some of the data during the training process, which harms data privacy to some extent. Recently, Federated Learning (FL) has attracted increasing attention for its efficiency in aggregating multiple local models trained from distributed data without privacy leakage. Yamamoto et al.\cite{Yamamoto} have applied FL to CPDP and propose a new method named FLR, which ensures privacy preservation.

Although the FL algorithm provides fairly satisfying privacy protection for traditional CPDP methods, it still does not take the data heterogeneity among multiple clients into account, which is one of the primary challenges within traditional CPDP. 
In practical applications, projects from different companies or organizations often have diverse data distributions due to variations in coding styles and business requirements \cite{xiaxin,Shriram}, which is often referred to as non-independent and identically distributed (Non-IID) data. In such a case, the local model from each client will update towards different optimization directions, leading to increased variance in the aggregated global model and the global model's performance degradation \cite{Sai,Xiang}.

In this paper, we propose a challenging and realistic problem, privacy-preserving CPDP with data heterogeneity. More specifically, we aim to deal with both performance with Non-IID data and privacy preservation in CPDP. In this setting, users should preserve the privacy of the local project data and collaboratively train a global model against data heterogeneity. It is more difficult than the single traditional CPDP or privacy-preserving CPDP. Furthermore, research by He et al. \cite{He} has demonstrated that carefully selected open-source projects can serve as valuable resources for building defect prediction models for proprietary projects in the centralized setting.
Based on this, a direct idea here to alleviate the data heterogeneity in CPDP with FL is to consolidate the model with additional training on open-source projects at the server. However, experiments show that a simple combination of open-source projects and CPDP with FL algorithms fails to produce promising results. The unstable training process arising from uneven open-source project data may worsen the performance of the global model.

To address this problem, we propose a novel privacy-preserving CPDP approach based on knowledge enhancement, named \textbf{FedDP}, that can effectively enhance the model performance with the knowledge from open-source projects by knowledge distillation. More specifically, FedDP follows the FL algorithms to protect the privacy of local projects and improves with two simple but effective solutions. Firstly, we apply the \textbf{Local Heterogeneity Awareness}, which holds that the local project data in each client constitutes a specific distribution, and identify the correlation factor between the local data and the open-source project data. Secondly, \textbf{Global Knowledge Distillation} will be done after the aggregation of local models at the server. We treat the aggregated global model as the student model and the weighted ensemble prediction of all local models as the teacher model. Knowledge transfer from teacher to student is achieved via knowledge distillation. Compared to simple parameter averaging aggregation methods, this knowledge transfer surpasses the mere parameter values, encompassing more extensively the deep-level information captured by local models, such as feature representations and decision boundaries. The open-source project data is no longer trained directly but serves as the distillation dataset, effectively mitigating the risk of exacerbating the degree of Non-IID that may arise from direct training. For a given distillation sample in the open-source project data, we endow the local model with high ensemble weight when its correlation factor is significant and vice versa. The principle behind this is the fact that a model tends to make the correct prediction when the sample fits the data distribution for training the model. With the proposed two solutions, our model significantly outperforms baselines on 14 projects from the Promise dataset and 5 projects from the Softlab dataset. The main contributions of this paper are as follows:

\begin{enumerate}
    \item We first solve the problem of privacy-preserving CPDP with data heterogeneity in real-world scenarios, where the distribution of project data among clients is different, and the privacy of each local project should be guaranteed. Subsequently, we propose open-source project-based knowledge enhancement for this challenge.
    \item We empirically find the unstable learning process with uneven open-source project data at the server can lead to notable performance degradation. Motivated by this observation, we further propose FedDP with local heterogeneity awareness and global knowledge distillation two solutions.
    \item We conduct extensive experiments on 19 projects from two datasets, and the results demonstrate that our method outperforms baselines by an average improvement of 2.66\% on the F1 metric.
\end{enumerate}

The remainder of this paper is structured as follows. Section \ref{back} presents background and related work, encompassing privacy-preserving CPDP, federated learning, and knowledge distillation. Section \ref{sec3} formulates the problem definition. Section \ref{meth} introduces a motivating experiment and our proposed approach, FedDP. Section \ref{ex_set} describes the experimental setup, while Section \ref{ex_result} validates the effectiveness of FedDP through thorough experimental results. Section \ref{dis} discusses the experimental details and the generalization capability of FedDP. Potential threats to the validity of this work are discussed in Section \ref{threat}. Finally, Section \ref{concl} concludes the paper and outlines directions for future research.

\section{Background and Related Work}\label{back}
% \li{too few references in this section}
\subsection{Privacy Preservation in CPDP}
CPDP addresses the problem of insufficient training data in the target project by leveraging data from other source projects \cite{back_cpdp1,back_cpdp2,back_cpdp3}. Depending on the number of available source projects, CPDP can be categorized into two types: Single-Source Cross-Project Defect Prediction (SSCPDP) and Multi-Source Cross-Project Defect Prediction (MSCPDP). When constructing CPDP models, SSCPDP assumes the availability of only one source project, whereas MSCPDP presumes the existence of multiple source projects\cite{master}. Numerous SSCPDP methods have been proposed in previous research. TCA+ \cite{tca+} integrates data normalization techniques with Transfer Component Analysis (TCA) to learn an optimal common feature subspace between the source and target projects. TNB \cite{tnb} assigns weights to training data instances by predicting the target project distribution, thereby constructing a weighted Naive Bayes classifier. In recent years, MSCPDP has garnered increasing attention from the academic community. Xia et al.'s Hybrid Model Reconstruction Approach (HYDRA) \cite{hydra}, which contains two phases: genetic algorithm and ensemble learning, offers a fresh perspective on MSCPDP. MASTER \cite{master} is a multi-source transfer weighted ensemble learning method that achieves three-fold transfer, i.e., dataset transfer, instance transfer, and feature transfer.

Despite the utility of CPDP, its foundation lies in data sharing, which poses privacy concerns and hinders its practical implementation. Notably, companies such as AT\&T have publicly expressed their reluctance to disclose their data \cite{Weyuker}. To overcome this limitation, many studies have started focusing on developing CPDP methods that incorporate privacy-preserving techniques. Peters et al. \cite{PetersLACE2} introduced LACE2, a privacy preservation strategy considering multi-party data. LACE2 effectively reduces the overall amount of data shared without compromising prediction performance. Li et al. \cite{Li_loc} developed a double obfuscation algorithm leveraging sparse representation and effectively implemented it within the context of CPDP. Yamamoto et al. \cite{Yamamoto} proposed FLR based on FL, a CPDP method that does not require data sharing and outperforms traditional privacy preservation methods. Yang et al. \cite{xiaxin} proposed Almity, a FL-based framework that addresses data security issues for software practitioners, thus bridging the gap between academic models and industrial applications.

However, these methods do not achieve better privacy protection or overlook the issue of data heterogeneity among various local projects in CPDP. Specifically, traditional methods, such as LACE2, still require disclosing a small amount of data. This requirement fundamentally limits their ability to achieve higher levels of privacy protection. CPDP methods based on FL do not adequately account for the data distribution differences among various projects. This Non-IID characteristic significantly impacts the accuracy and effectiveness of the models.

\subsection{Federated Learning}
Federated learning (FL) has emerged as a new machine learning paradigm to allow distributed training for clients \cite{wang2023fedcda,li2024unleashing,li2024pfeddil}. Unlike traditional centralized methods that collect and aggregate data on a server or cluster for training \cite{zhu}, FL preserves privacy and security by allowing clients to train the data locally \cite{yang,Li_2024_CVPR,li_tpds}. 
One effective architecture for FL is FedAvg \cite{fedavg}. After several epochs of local updates, FedAvg performed the weighted averaging of the models uploaded by clients on the central server, with the weights being determined by the size of local data on each client. However, traditional FL algorithms like FedAvg suffer data heterogeneity, where the datasets in clients are
Non-IID, resulting in degradation in model performance \cite{back_fl1,backfl2,li2024rehearsal}. To address the Non-IID challenge, FedProx \cite{FedProx} introduced a proximal term into the objective function of local models to prevent them from deviating too far from the global model. SCAFFOLD \cite{Sai} introduced an additional parameter, known as the control variate, to correct the direction of local updates. 

Despite the numerous methods proposed in FL to alleviate the challenges posed by data heterogeneity, applying them to other downstream tasks like CPDP remains in many trials. These parameter averaging aggregation approaches have limited effectiveness in mitigating the Non-IID issue.

\subsection{Knowledge Distillation}
Knowledge distillation (KD) is the process of transferring knowledge from one or more teacher models to another student model \cite{KDLOGIT1,KD2}. The key to KD is to align the soft predictions of the student model with the soft predictions of the teacher model \cite{KD1,KD3}. 
Currently, KD has been applied in FL to address the challenges posed by Non-IID data by extracting knowledge from multiple local models and distilling it into the global model \cite{fkd1,fkd2}. FedDF \cite{FKDDF} is a federated knowledge distillation approach for model fusion, which utilizes a public dataset as the distillation data to extract knowledge from multiple local models into a global model. However, considering that such auxiliary datasets may be lacking in real-world environments, FedGEN \cite{fedgen} learns a lightweight generator to perform federated knowledge distillation. DaFKD \cite{dafkd} generates distillation samples through the generator and performs weighted distillation by determining the weights between the samples and the corresponding domain according to the domain discriminator.
% DaFKD \cite{dafkd} further considers model diversity by assigning weights based on the importance of local models and then distilling them after the weighted ensemble.

The successful application of knowledge distillation in FL offers a feasible solution for addressing data heterogeneity and privacy concerns in CPDP. However, due to significant differences among various research domains, such as data characteristics and model architectures, many FL solutions tailored for Non-IID problems cannot be directly transferred for other tasks.

\section{Problem Formulation}\label{sec3}
\begin{figure}[t] 
    \centering
    \includegraphics[width=0.7\linewidth]{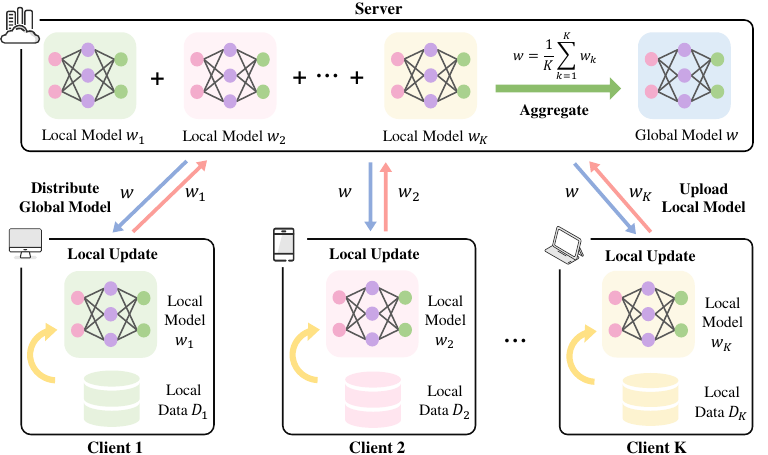}
    % \Description{In FL, clients first update local models based on the distributed global model. Then, the updated local models are uploaded to the server and aggregated to obtain a global model for the next communication round.}
    \caption{Overview of Federated Learning. Clients first update local models based on the distributed global model. Then, the updated local models are uploaded to the server and aggregated to obtain a global model for the next communication round.} 
    \label{fig1}
\end{figure}
A typical FL problem can be formalized by collaboratively training a global model for $K$ total clients in FL. We consider each client $k$ can only access to his local private dataset $D_k=\{x_i^k, y_i^k\}$, where $x_i^k$ is the $i$-th input data sample and $y_i^k \in \{1,2,\cdots,C\}$ is the corresponding label of $x_i^k$ with $C$ classes. We denote the number of data samples in dataset $D_k$ by $|D_k|$. The global dataset is considered as the composition of all local datasets $D=\{D_1, D_2,\cdots, D_K\}$, $D=\sum_{k=1}^{K}D_k$. 

The FL framework is illustrated in Fig. \ref{fig1} and involves the following steps: 
\begin{enumerate}[(1)]
    \item The central server distributes the initial unlearned model to each client.
    \item After receiving the distributed model from the server, clients participating in the training process train models using their own local data.  
    \item The participating clients upload the model parameters to the central server.
    \item The central server aggregates the uploaded local models to update the global model and sends it to participating clients in the next communication round.
\end{enumerate}

Steps (2) to (4) constitute a complete communication round. By repeating this cycle, the clients and central server continuously communicate until the global model converges. The objective of the FL learning system is to learn a global model $w$ that minimizes the total empirical loss over the entire dataset $D$:
\begin{align}
    &\min_{w} \mathcal{L}(w):= \sum_{k=1}^K \frac{|D_k|}{|D|}\mathcal{L}_k(w), \nonumber \\ 
    &\text{where} \ \mathcal{L}_k(w) = \frac{1}{|D_k|} \sum_{i=1}^{|D_k|} \mathcal{L}_{CE}(w; x_i^k, y_i^k),
    \label{eq:conventional_loss}
\end{align}
where $\mathcal{L}_k(w)$ is the local loss in the $k$-th client and $\mathcal{L}_{CE}$ is the cross-entropy loss function that measures the difference between the prediction and the ground truth labels.

In the privacy-preserving cross-project defect prediction task, each client denotes each version of a project in the studied datasets, and the object of each client is to predict defects within the local project, which is a binary classification task. Moreover, open-source projects are abundant and available in real-world scenarios and here we denote the open-source project data at the server as $\mathbb{D} = \{x_i^d, y_i^d\}$, where $x_i^d$ is the $i$-th data sample and $y_i^d \in \{1,2,\cdots, C\}$ is the corresponding label of $x_i^d$ with $C$ classes. The detailed employment of open-source project data will be discussed in the following Section \ref{meth}.
\section{Methodology}\label{meth}
In this section, we first discuss an interesting observation, i.e., simply combining open-source projects and CPDP with FL algorithms does not work well. Motivated by the finding, we finally present our FedDP framework.

\subsection{Why does intuitive combination fail to work well in privacy-preserving CPDP?}\label{openflr_analysis}

\begin{figure}[h] 
    \centering
    % \Description{In the OpenFLR, the performance experiences undesired fluctuation in F1 and AUC two metrics with different open-source project data.}
    \includegraphics[width=0.7\linewidth]{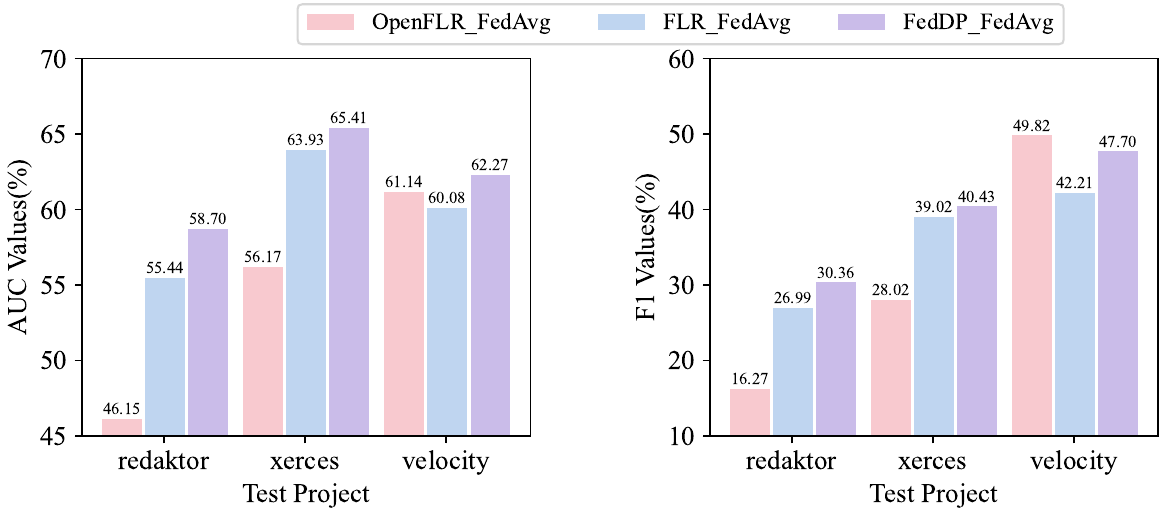}
    \caption{Comparison of performance on three different open-source projects in terms of AUC and F1 values.} 
    \label{fig5}
\end{figure}

After receiving the updated models with Eq.(\ref{eq:conventional_loss}) from participating clients at each communication round $t$, the server first aggregates the models to obtain the global model $w$:

\begin{equation}\label{aggregated_model}
\begin{split}
 w_{t+1} = {\sum_{k\in S_{t}}p_k \cdot w_t^k},\\
\text{where} \ p_k = \frac{|D_k|}{\sum_{i \in S_t}|D_i|}
\end{split}
\end{equation}
where $\left\vert S_t\right\vert = K \times R$ and $R$ denotes client participation ratio. Then, we utilize the open-source project data $\mathbb{D}$ as the training data and continually train the aggregated global model $w_{t+1}$ with the cross-entropy loss, which is defined in Eq.(\ref{eq:conventional_loss}). Here we denote the global model after training with the open-source project data as $\hat{w_{t+1}}$, and the illustration of the experiment results is shown in Fig. \ref{fig5}.

Unfortunately, simply combining open-source projects and CPDP with FL algorithms yields unpromising results. We report three methods with two common metrics on three test projects here, which will be detail described in Section \ref{ex_result} later. FLR represents the basic CPDP with FL algorithms while OpenFLR is the model with additional training on the open-source project data. In the OpenFLR, the performance experiences undesired fluctuation in F1 and AUC two metrics with different open-source project data. The explanation behind this is the local model from each client will update towards different optimization directions due to the Non-IID data, and the unknown shift distribution between local project data and open-source project data may worsen the model performance. 
To ensure the effectiveness of knowledge enhancement using open-source projects regardless of diverse data distributions, we developed FedDP (ours) and will discuss it in the next section, which also demonstrates a significant performance in Fig.\ref{fig5}.

\begin{figure*}[t]
    \centering
    \includegraphics[width=\linewidth]{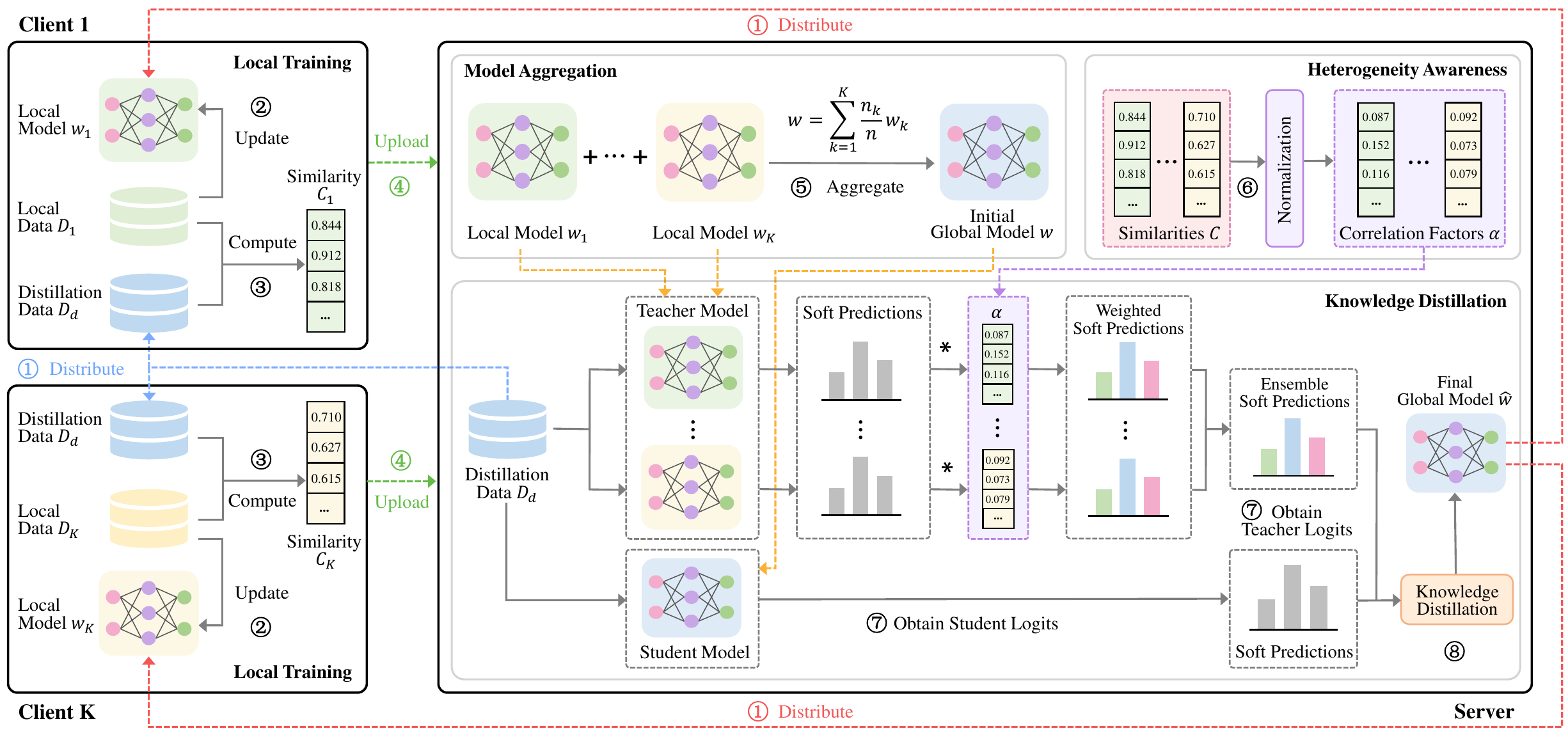}
    % \Description{The server distributes the model and distillation data to each client (step 1). The client then trains its local model and identifies correlation factors between the local data and open-source data (steps 2-3). Clients upload the local model and the correlation factors to the server (step 4). The server aggregates the local models and performs knowledge distillation with correlation factors (steps 5-8).}
    \caption{Overview of FedDP between a central server and clients. The server distributes the model and distillation data to each client (step 1). The client then trains its local model and identifies correlation factors between the local data and open-source data (steps 2-3). Clients upload the local model and the correlation factors to the server (step 4). The server aggregates the local models and performs knowledge distillation with correlation factors (steps 5-8).}
    \label{fig2}
\end{figure*}

\subsection{FedDP}
The key object of FedDP is to harness open-source data for knowledge enhancement, thus alleviating the data heterogeneity problem caused by Non-IID project data in privacy-preserving CPDP. Specifically, each client in FedDP first trains a local model using its local private data and discerns the correlation factor of local project data to the open-source project data. Then, the server conducts knowledge distillation on the open-source project data to enhance the knowledge of the global model. During this process, Here we build the student model with the aggregated global model, whereas the weighted ensemble prediction with correlation factors derived from all local models functions as the teacher model. 
% Subsequently, the central server assigns personalized weights to the clients based on their learning proficiency about open-source information, reflecting the trustworthiness of each local model. The central server then performs a weighted ensemble of the local models according to the personalized weights to construct a teacher model. Concurrently, the central server considers the aggregated global model as the student model and employs the open-source projects as the distillation dataset. Knowledge distillation is then conducted through soft predictions by the student and teacher models on the distillation dataset. 
Fig. \ref{fig2} illustrates an overview of the training framework of FedDP between a central server and clients. The specific steps of the FedDP algorithm are outlined in Algorithm \ref{algorithm}.

\subsubsection{Local Heterogeneity Awareness}
In the global knowledge distillation, we obtain the teacher model by ensemble soft predictions of uploaded models. To endow the model with suitable weight to the distillation dataset, an intuition is that the model has a high probability of making the correct prediction when the sample fits in the training data distribution. As a consequence, quantifying the correlation between the local data and the open-source project data is necessary. On the client side, each client receives the open-source project data (distillation data) from the central server after updating the local model. Then, we introduce the local heterogeneity awareness that clients employ the cosine similarity function to compute the correlation factor between its local data and the distillation data. 
Supposed that the client $k$ has finished the model updating on the local data, we compute the cosine similarity $C_{i}^k$ between each distillation sample $x_i^d \in \mathbb{D}$ and all samples of local data $D_k$, and then take the average to represent the similarity between the distillation samples $x_i^d$ and $D_k$:
\begin{equation}
\label{eq3}
C_i^k = \frac{1}{|D_k|} \sum_{j=1}^{|D_k|}Cosine\_similarity(x_i^d, x_j^k)
\end{equation}
where $x_j^k$ is the $j$-th data sample from the local data $D_k$ and $Cosine\_similarity(\cdot)$ represents the function used to compute the cosine similarity between data samples. Each client will upload both its updated local model and the correlation factor to the server. Upon completion of uploads from all participating clients, the central server will perform global knowledge distillation.

\begin{algorithm}[t]
\caption{FedDP}
\label{algorithm}
\begin{algorithmic}[1] 
\Require 
	$\mathbb{D}$:distillation dataset; $D_k$:local dataset;  $E$: local training epoch; 
 
 \hspace{0.4em}$N$: distillation step;    $K$: client number; $T$: communication round;    
 
 \hspace{0.4em}$R$:  client participation ratio; $\eta$: learning rate; 
 
 \hspace{0.4em}$p$: distillation data sampling size.
\Ensure 
$w$: global model.

\hspace{-3.5em}\underline{\textbf{Server executes:}}

\State Initialize global model ${w}_1$ 

\For{$t=1,2,\cdots, T$}

    \State $m \leftarrow \max(R\cdot K, 1)$
    
    \State ${S}_{t},{S}_{t}^{d} \leftarrow$ randomly select $m$ clients and $p$ distillation samples 
    % from $\mathbb{D}$
    % \State  $S_{t}, S_{t}^{d}\leftarrow \text{randomly select subsets of clients and distillation samples based on} R \text{and} p$ 
    
     \For{each selected client $k \in S_t$ {\rm\bfseries in parallel}}
     
     \State   $w^k_t, C^{k} \leftarrow \text{ClientUpdate}(k, {w}_{t})$ 

    \EndFor
    
    \State ${\alpha}^{k} \leftarrow$ obtain personalized weights with (\ref{normalization}) 
        
    \State${w}_{t+1} \leftarrow$ aggregate local models with (\ref{aggregated_model}) 
    
    \For{$j=1,2,\cdots, N$}
    
    \State input ${S}_{t}^{d}$ to ${w}_{t+1}$ and $w^k_t$ to obtain soft predictions 
    
   \State $\hat{w}_{t+1} \leftarrow$ update global model ${w}_{t+1}$ with (\ref{distillation_loss}) 

    \EndFor
\EndFor

\Statex

\hspace{-1.5em}\underline{\textbf{Client update$(k,{w}_{t})$:}}

 \State receive initial model ${w}_{t} $  from server 

\For{$e=1,2,\cdots,E$}

\State ${w}_{t}^{k} \leftarrow $ update local model ${w}_{t}$ with (\ref{eq:conventional_loss}) 

\EndFor
\State ${C}^{k} \leftarrow$ obtain cosine similarity with (\ref{eq3}) 

\State \textbf{return} $w_{t}^{k},{C}^{k}$ to server
\end{algorithmic}
\end{algorithm}

% \subsection{Federated Knowledge Distillation Process on the Central Server}
  
\subsubsection{Global Knowledge Distillation}
After receiving the local models and correlation factors from each participating client in $S_t$ at the communication $t$, the central server first aggregates the local models to obtain the global model $w_{t+1}$ with Eq.(\ref{aggregated_model}). Motivated by the knowledge distillation techniques \cite{method_kd2,KD1}, which encourage the student model to approximate the output logits of the teacher model, the student is able to imitate the teacher’s behavior with the loss on the distillation dataset. To alleviate the optimization variances in the global model, in this paper, we treat the open-source project data as the distillation dataset and distill the knowledge from the teacher model (built with the ensemble predictions of uploaded local models) to the student model (built with the global model).
% , aiming to alleviate the shift of optimization directions in the global model. 
% In contrast, ensemble learning methods allow to combine multiple heterogeneous weak classifiers by averaging the predictions of the individual models to improve the model performance. 
% We build the student model with the ensemble predictions of uploaded local models.
% the To better enhance the performance of the global model with the knowledge of open-source project data, we 
% In contrast, we observe that a model tends to make the correct prediction when the distillation sample fits the training data distribution.
Then, we assign each local model with personalized weights according to the correlation factor, which reflects the similarity between local data and distillation data. For each distillation sample $x_i^d \in \mathbb{D}$, the central server normalizes correlation factors from all participating clients to compute the personalized weight $\alpha^k_i$ for each client $k$:
\begin{equation}\label{normalization}
\alpha_i^k = \frac{C_i^k}{\sum_{k\in S_{t}} C_i^k}\ ,\ \ \sum_{k\in S_{t}}\alpha_i^k = 1
\end{equation}
Here a higher similarity indicates that the distribution of the client's local data is more similar to the distillation data.

Subsequently, an ensemble distillation process is performed to train the aggregated global model $w_{t+1}$. We input the distillation samples ${x_i^d \in \mathbb{D}}$ into both the global model $w_{t+1}$ and each local model $w^k_t$ to obtain the soft predictions $s(w_{t+1}; x_i^d)$ and $s(w^k_t;x_i^d)$. After that, we perform a weighted ensemble of the soft predictions from the local models based on personalized weights, serving as the teacher model's logits. In the next step, we utilize Kullback-Leibler divergence (KL-divergence) \cite{Kl} as the distillation loss, enabling the student model to learn the knowledge from the teacher model. By minimizing the distillation loss between the logits of the teacher model and the logits of the student model, we can gradually update the student model and obtain the final global model $\hat{w}_{t+1}$:
\begin{equation}\label{distillation_loss}
    \begin{aligned}
        \hat{w}_{t+1}&=\arg \min_{w_{t+1}} \mathcal{L}_{KD}(w_{t+1})  \\
        &= \frac{1}{|\mathbb{D}|}\sum_{x_i^d \in \mathbb{D}}\! \mathop{K\!L}\big(\sum_{k=1}^{S_t} \alpha_i^k \cdot s(w^k_t; x_i),s(w_{t+1}; x_i^d)\big) 
% &\hat{w}_{t,j} = \hat{w}_{t,j-1} - \eta \frac{\partial KL( f(w^k_t; S_t^d), f(\hat{w}_{t,j-1}; S_t^d) )}{\partial \hat{w}_{t,j-1}}  \\
% & \quad \text{where }  f(w^k_t; S_t^d) = \sigma(\sum_{k \in S_t} \sum_{x_i \in S_t^d}\alpha_{k,i} \cdot s(w^k_t;x_i^d)) \\
% &\quad \quad \quad f(\hat{w}_{t,j-1}; S_t^d) = \sigma(\sum_{x_i \in S_t^d} s(\hat{w}_{t,j-1}; x_i^d))
    \end{aligned}
\end{equation}
where $KL(\cdot)$ is to compute the KL-divergence.After $N$ epochs of knowledge distillation, we will obtain a new global model $\hat{w}_{t+1}$. In the next communication round, the server distributes the global model $\hat{w}_{t+1}$ to each participating client for local training.
% 调整篇幅
% After performing $N$ epochs of knowledge distillation, we will obtain a new global model $\hat{w}_{t+1}$. In the next communication round, the server distributes the global model $\hat{w}_{t+1}$ to each participating client for local training.

\section{Experimental Setup}\label{ex_set}
\subsection{Dataset}\label{dataset}

\begin{table*}[ht]
\caption{Details of Software Projects used in experiments.\label{tab:table0}}
\centering
% \begin{tabular*}{0.7\textwidth}{@{\extracolsep\fill}cclccc@{}}
\begin{tabular}{c|c|c|l|cccc}
\bottomrule
Dataset&Language&Project  &Description & Versions & Instances & Defective Rate (\%) & Category \\ \hline
\multirow{26}{*}{Pomise}&\multirow{26}{*}{Java}&\multirow{2}{*}{ant}  & \multirow{2}{*}{A build management system} & 1.6& 351 & 26.21 & MM\\
& & & & 1.7 & 745 & 22.28 &HM \\
\cline{3-8}
% &camel &A versatile integration framework & 1.4,1.6 & 872-965 & 16.63-19.48 & \\
 & &\multirow{2}{*}{camel}&\multirow{2}{*}{A versatile integration framework} & 1.4 & 872 & 16.63 & $\backslash$ \\
 & & & & 1.6 & 965 & 19.48 &  $\backslash$ \\
\cline{3-8}
% &jedit &A text editor & 4.0,4.1 & 306-312 & 24.51-25.32& \\
 & &\multirow{2}{*}{jedit}&\multirow{2}{*}{A text editor} & 4.0 & 306 & 24.51 &MM \\  
 & & & & 4.1 & 312 & 25.32 &MM \\ \cline{3-8}
% &lucene & A text search engine library & 2.2,2.4 & 247-340 & 58.30-59.70& \\
 & &\multirow{2}{*}{lucene}&\multirow{2}{*}{A text search engine library} & 2.2 & 247 & 58.30 &MH \\  
 & & & & 2.4 & 340 & 59.70 & MH\\ \cline{3-8}
% &xerces & An XML processor & 1.2,1.3 & 440-453 & 15.23-16.14& \\
 & &\multirow{2}{*}{xerces}&\multirow{2}{*}{An XML processor} & 1.2 & 440 & 16.14 & ML\\  
 & & & & 1.3 & 453 & 15.23 &ML \\ \cline{3-8}
% &velocity & A template language engine  & 1.5,1.6 & 214-229 & 34.06-66.35& \\
 & &\multirow{2}{*}{velocity}&\multirow{2}{*}{A template language engine} & 1.5 & 214 & 66.35 &MH \\  
 & & & & 1.6 & 229 & 34.06 &MH \\ \cline{3-8}
% &xalan  & An XSLT processor & 2.5,2.6 & 803-885 & 46.44-48.19& \\
 & &\multirow{2}{*}{xalan}&\multirow{2}{*}{An XSLT processor} & 2.5 & 803 & 48.19 &HH \\  
 & & & & 2.6 & 885 & 46.44  &HH \\ \cline{3-8}
% &synapse & A high-performance Enterprise Service Bus & 1.1,1.2 & 222-256 & 27.03-33.59& \\
 & &\multirow{2}{*}{synapse}&\multirow{2}{*}{A high-performance Enterprise Service Bus} & 1.1 & 222 & 27.03 & MM\\  
 & & & & 1.2 & 256 & 33.59 & MH\\ \cline{3-8}
% &log4j & A logging utility& 1.0,1.1 & 109-135 & 25.18-33.94& \\
 & &\multirow{2}{*}{log4j}&\multirow{2}{*}{A logging utility} & 1.0 & 135 & 25.18 &LM \\  
 & & & & 1.1 & 109 & 33.94 & LH\\ \cline{3-8}
% &poi & API for Office Open XML standards& 2.5,3.0 & 385-442 & 63.57-64.41& \\
 & &\multirow{2}{*}{poi}&\multirow{2}{*}{API for Office Open XML standards} & 2.5 & 385 & 64.41 &MH \\  
 & & & & 3.0 & 442 &  63.57& MH\\ \cline{3-8}
% &ivy & A dependency manager & 1.4,2.0 & 241-352 & 6.64-11.36& \\
 & &\multirow{2}{*}{ivy}&\multirow{2}{*}{A dependency manager} & 1.4 & 241 & 6.64 &ML \\  
 & & & & 2.0 & 352 & 11.36 &ML \\ \cline{3-8}
 & &prop6 & An industrial projects in the insurance domain &  & 660 & 10.00& HL\\ \cline{3-8}
 & &redaktor & A decentralized content management system &  & 176 & 15.34& LL\\ \cline{3-8}
 & &tomcat & A Web server&  & 858 & 8.97& HL\\  \cline{3-8}
\cline{1-7}
% \multirow{5}{*}{Softlab}&ar1  & Embedded software &  & 121 & 7.44 \\
% &ar3  & Dishwasher &  & 63 & 12.70 \\
% &ar4  & Refrigerator &  & 107 & 18.69 \\
% &ar5  & Washing machine &  & 36 &22.22 \\
% &ar6  & Embedded software &  & 101 &14.85 \\
\multirow{5}{*}{Softlab} &  \multirow{5}{*}{C}&
ar1 & Embedded software & & 121 & 7.44 & $\backslash$ \\   \cline{3-8}
& & ar3 & Dishwasher & & 63 & 12.70&ML \\   \cline{3-8}
& & ar4 & Refrigerator & & 107 & 18.69 &MM\\   \cline{3-8}
& & ar5 & Washing machine & & 36 & 22.22&LM \\   \cline{3-8}
& & ar6 & Embedded software & & 101 & 14.85&ML \\   \cline{3-8}
\toprule
\end{tabular}
\end{table*}

\begin{table*}[ht]
\centering
\caption{Categorization of Data Distributions.\label{tab:add1}}
\begin{tabular}{l|c|c|c}
\bottomrule
% \diagbox{Scale}{Balance}& Low(L) & Medium(M) & High(H) \\
% \hline
% Low(L)     & LL     & LM        & LH \\
% Medium(M)  & ML     & MM        & MH  \\
% High(H)    & HL     & HM        & HH  \\
\diagbox{Scale}{Balance} & Low (L) & Medium (M) & High (H) \\
\hline
Low (L)     & LL     & LM        & LH \\
Medium (M)  & ML     & MM        & MH  \\
High (H)    & HL     & HM        & HH  \\
\toprule
\end{tabular}
\end{table*}

In our experiments, we utilize the Promise\footnote{https://github.com/opensciences/opensciences.github.io} and Softlab\footnote{https://zenodo.org/record/1209483} datasets, which are widely employed in most defect prediction research \cite{Li_loc,promise1,promise2}. Promise includes multiple projects developed in JAVA. We filter out smaller projects based on the number of instances. Finally, we select 14 projects with 25 versions for our experiments and choose the project `camel' with the maximum instance count as the open-source project. Softlab contains 5 C-language projects, and we similarly select the largest project, `ar1', as the open-source project.
Table \ref{tab:table0} provides specific information about the dataset, including the project description, the project version numbers, the instance counts, the ratio of defect instances, and the category of data distribution.

Drawing upon the research conducted by Yang et al. \cite{xiaxin}, we evaluate the degree of Non-IID in CPDP data across two dimensions: data scale and data balance. For each dimension, we establish two thresholds to categorize them into three levels: Low (L), Medium (M), and High (H). Specifically, the data scale is classified based on the proportion of instance count to its ideal value, where less than 50\% is designated as L, between 50\% and 150\% as M, and above 150\% as H. The ideal value is calculated as the total number of training instances divided by the number of clients. The data balance is categorized using thresholds of 16.67\% (representing a 1:5 ratio of minority to majority classes) and 33.33\% (representing a 1:2 ratio). The data distribution can be segmented into nine categories by combining these two dimensions, as illustrated in Table \ref{tab:add1}.

To illustrate with an example, consider a FL system consisting of 500 training data and 5 clients. Ideally, each client should have 100 instances and a 50\% data balance ratio (1:1 minority-to-majority class ratio). Suppose a client possesses only 28 instances (representing 28\% of the ideal value) and exhibits a defect rate of 25\% (indicating a 1:3 ratio of minority to majority classes). In that case, the data distribution of this client can be classified as LM. Table \ref{tab:table0} shows the distribution for the Promise and Softlab datasets excluding distillation data (i.e., project `camel' and project `redaktor'). Promise covers all types and Softlab covers three.

\subsection{Evaluation}
\subsubsection{Evaluation Metrics}

In the task of CPDP, metrics such as Precision, Recall, F1 Score, and AUC are frequently employed to evaluate the predictive performance of models \cite{metric1,metric2,metric3}. Additionally, in FL, the number of communication rounds is typically used as a metric to assess communication efficiency \cite{FKDDF,fedavg}. Next, we will specifically introduce the definitions of these metrics.

% \noindent \textbf{Metrics for Evaluating Predictive Performance:}
\begin{itemize}
\item{\textbf{Precision:} It is the proportion of true positive predictions among all instances predicted as positive by the model and represents the degree of classification accuracy of the model. The detailed definitions are shown as follows:}
\begin{equation}
Precision = \frac{TP}{TP + FP} 
\end{equation}
where $TP$ refers to the number of instances where the model predicts a positive sample and the actual sample is also positive, and $FP$ refers to the number of instances where the model predicts a positive sample but the actual sample is negative.
\item{\textbf{Recall: } It is a crucial metric that quantifies the model's ability to capture all true positive examples. In the task of CPDP, a high Recall value indicates that the model can identify a larger number of potential defects, which is extremely important for software quality assurance, as neglected defects could lead to software failures or security breaches. The detailed definitions are shown as follows:}
\begin{equation}
Recall = \frac{TP}{TP + FN} 
\end{equation}
where $FN$ refers to the number of instances where the model predicts a negative sample but the actual sample is positive.
\item{\textbf{F1:} It is the harmonic mean of Precision and Recall, used to achieve a balance between the two. Particularly in the face of imbalanced data, the F1 can provide a balanced performance evaluation.}
\begin{equation}
F1 = 2 \times \frac{Precision \times Recall}{Precision + Recall} 
\end{equation}
\item{\textbf{AUC:} It is a commonly used metric in binary classification tasks and is also widely applied in CPDP. It measures the overall performance of a model by evaluating the true positive rate and false positive rate across different classification thresholds. The AUC value ranges from 0 to 1, with 0.5 denoting random predictions. A higher AUC value indicates better classification capability.}
\end{itemize}

% \noindent \textbf{Metrics for Evaluating Communication Efficiency:}
\begin{itemize}
\item{\textbf{Communication Rounds:} In FL, communication efficiency is typically assessed by the number of communication rounds required to achieve the target performance. Each round refers to the transmission of models or gradients between all participating clients and the server. Unlike centralized learning, FL necessitates the exchange of model parameters between nodes, and these communication processes are often the most time-consuming part of the entire FL system. Therefore, reducing the number of communication rounds can significantly improve the efficiency of FL and reduce communication costs.}
\end{itemize}

\subsubsection{Significance analysis}
We employ the Win/Tie/Loss metric alongside the Wilcoxon signed-rank test to compare the performance differences between FedDP and baseline models. Both evaluation methods have been extensively adopted and applied within the relevant field \cite{xztest1,Li_loc}. Specifically, the Wilcoxon signed-rank test, as a non-parametric statistical approach, is utilized to assess the differences between two matched samples. By calculating the p-value, we can determine whether the difference between these two matched samples reaches a statistically significant level (i.e., p-value < 0.05).

Based on the p-value, we further categorize the outcomes into Win, Tie, or Loss: for each comparison between a baseline model M and FedDP, if the p-value is less than 0.05 and FedDP outperforms M, it is classified as a Win; if the p-value is less than 0.05 and M performs better, it is classified as a Loss; if neither condition is met, it is classified as a Tie.

\subsection{Baseline}

To assess the effectiveness of our proposed method, we compare it with  FLR \cite{Yamamoto}, Almity \cite{xiaxin}, and OpenFLR. These baselines can represent the state-of-the-art in this field. Additionally, we consider the classic FL algorithms FedAvg \cite{fedavg} and FedProx \cite{FedProx} to explore the potential impact of different FL algorithms. Here is a brief introduction to the FL algorithms and baseline models.

\noindent \textbf{FL Algorithms:}
\begin{itemize}
    \item FedAvg \cite{fedavg}: A FL algorithm that aggregates a global model by weighted averaging the local model parameters based on the size of the local data across different clients. 
    \item FedProx \cite{FedProx}: An extension to FedAvg, FedProx introduces a proximal term to the optimization objective, which helps in controlling the extent of divergence of the local models from the global model. 
\end{itemize}

\noindent \textbf{Baseline Models:}
\begin{itemize}
    \item Centralized Training: A traditional machine learning training approach where all training data is pooled together to train a classifier for defect prediction. It is often regarded as the upper bound for the performance of FL.
    \item FLR \cite{Yamamoto}: A privacy-preserving method for CPDP that employs logistic regression as its learning model and FedAvg as the algorithm for aggregating model parameters.
    \item OpenFLR: On the basis of FLR, open-source data is added to the training set. After aggregating the local models, the server further utilizes the open-source data to train the global model.

    \item Almity \cite{xiaxin}: An FL framework tailored for SE tasks, which introduces an innovative parameter aggregation strategy. This strategy considers three attributes: data scale, data balance, and minority class learnability.
\end{itemize}

\subsection{Configuration}

For the Promise dataset, excluding the project `camel' as the open-source project, we consider the latest version of each remaining project as the test data, resulting in a total of 13 test datasets. The treatment for the project `camel' is as follows: among all methods, only FLR's training process does not use the data of `camel'. In other words, the project `camel' serves as a training dataset in Centralized Training and OpenFLR, while it serves as the distillation dataset in FedDP. Additionally, our task is CPDP. When the latest version of a project is selected as test data, the remaining versions of that project are not included in the training. All versions of the remaining projects are used in training. Furthermore, we perform random oversampling on all training data to alleviate the impact of imbalanced data, but we do not process the distillation and test data. We applied the same configuration to the Softlab dataset.

In our experiments, each version of the participating projects acts as a separate client, and the global model employs a logistic regression model. We set the number of local training epochs $E = 10$, communication rounds $T = 50$, distillation steps $N = 10$, distillation data sampling size $p = 700$, client participation ratio $R = 100\%$, and a local training learning rate of 0.001. For each test data, we take the average of the last 10 communication rounds as the result of one experiment; we repeat such experiments 5 times and calculate the average performance.

\subsection{Research Questions}
This paper answers the following three research questions:
\begin{itemize}
\item RQ1: How does the predictive performance of FedDP compare to that of other privacy-preserving CPDP methods?

\item RQ2: How does the communication efficiency of FedDP compare to that of other FL-based CPDP methods?

\item RQ3: Is our proposed FedDP sensitive to different distillation datasets?
\end{itemize}

\section{Experimental Result}
\label{ex_result}
\subsection{RQ1: How does the predictive performance of FedDP compare to that of other privacy-preserving CPDP methods?}

\begin{table*}[!h]
\caption{Comparison of FedDP and baseline methods in terms of Precision value.\label{tab:table1}}
\centering
\resizebox{\columnwidth}{!}{
\begin{tabular}{cccccccc>{\columncolor{blue!10} }c>{\columncolor{blue!10} }ccc}
% \begin{tabular*}{\textwidth}{@{\extracolsep\fill}c|c|ccccc|cc@{}}
\bottomrule
\rowcolor{white}
  \multirow{3}{*}{Project}& Precision &  &  &  &  &  &  & &  &$p$-value&\\ 
\cmidrule(r){2-10} \cmidrule(r){11-12}
 & LR & OpenFLR & OpenFLR & FLR & FLR &Almity &Almity & \cellcolor{white}{FedDP} & \cellcolor{white}{FedDP} & vs. FLR &vs. Almity\\  

 & Centralized &FedAvg & FedProx & FedAvg & FedProx & FedAvg & FedProx& \cellcolor{white}FedAvg & \cellcolor{white}FedProx &FedProx &FedProx\\  
\hline
 ant & 43.53$_{\pm 0.012}$ & 33.53$_{\pm 0.006}$ & 33.65$_{\pm 0.004}$ & {48.06}$_{\pm 0.012}$ & \underline{49.43}$_{\pm 0.008}$ & \textbf{50.57}$_{\pm 0.009}$ & 46.80$_{\pm 0.009}$ &45.27$_{\pm 0.010}$ & 45.49$_{\pm 0.012}$ & <0.05 & <0.05\\
 jedit    & 44.74$_{\pm 0.008}$ & 35.85$_{\pm 0.003}$ & 35.83$_{\pm 0.006}$ & 49.12$_{\pm 0.026}$ & 47.39$_{\pm 0.018}$ & \textbf{55.20}$_{\pm 0.016}$ & 49.65$_{\pm 0.018}$ & {49.39}$_{\pm 0.010}$ & \underline{49.66}$_{\pm 0.007}$ & <0.05 & 0.972\\
 lucene   & 75.31$_{\pm 0.014}$ & 70.56$_{\pm 0.009}$ & 70.43$_{\pm 0.011}$ & 76.45$_{\pm 0.022}$ & 77.00$_{\pm 0.013}$ & 74.21$_{\pm 0.009}$ & 72.13$_{\pm 0.006}$ &   \textbf{77.73}$_{\pm 0.013}$ & \underline{77.07}$_{\pm 0.013}$ & 0.865 & <0.05\\

xerces   & \textbf{41.70}$_{\pm 0.011}$ & 19.33$_{\pm 0.006}$ & 19.30$_{\pm 0.007}$ & \underline{38.44}$_{\pm 0.039}$ & 37.80$_{\pm 0.009}$ & 35.57$_{\pm 0.011}$ & 32.59$_{\pm 0.012}$ & 36.19$_{\pm 0.008}$ & 36.29$_{\pm 0.009}$ & <0.05 & <0.05 \\  
velocity & 53.40$_{\pm 0.021}$ & 46.94$_{\pm 0.004}$ & 47.42$_{\pm 0.003}$ & \textbf{56.16}$_{\pm 0.014}$ & \underline{55.79}$_{\pm 0.018}$ & 54.02$_{\pm 0.018}$ & 52.05$_{\pm 0.011}$ & 55.54$_{\pm 0.011}$ & 55.05$_{\pm 0.011}$ & 0.448 & <0.05 \\  
xalan    & 55.36$_{\pm 0.009}$  & {63.07}$_{\pm 0.007}$ & {63.33}$_{\pm 0.004}$ & 56.87$_{\pm 0.006}$ & 55.73$_{\pm 0.012}$ & \textbf{69.16}$_{\pm 0.011}$ & \underline{66.06}$_{\pm 0.016}$ & 57.26$_{\pm 0.002}$ & 57.12$_{\pm 0.002}$ & <0.05 & <0.05 \\  
synapse  & 62.34$_{\pm 0.021}$ & 45.41$_{\pm 0.004}$ & 45.08$_{\pm 0.003}$ & {65.00}$_{\pm 0.016}$ & {65.28}$_{\pm 0.014}$ & \textbf{69.88}$_{\pm 0.011}$ & \underline{66.74}$_{\pm 0.031}$ & 59.94$_{\pm 0.014}$ & 60.61$_{\pm 0.007}$ & <0.05 & <0.05 \\  
log4j    & 71.02$_{\pm 0.020}$ & 42.25$_{\pm 0.010}$ & 40.88$_{\pm 0.017}$ & 73.17$_{\pm 0.048}$ & 72.36$_{\pm 0.032}$ & 58.51$_{\pm 0.053}$ & 53.83$_{\pm 0.010}$ & \textbf{80.46}$_{\pm 0.018}$ & \underline{79.94}$_{\pm 0.017}$ & <0.05 & <0.05 \\  
poi      & 84.85$_{\pm 0.005}$ & 75.31$_{\pm 0.007}$ & 75.15$_{\pm 0.006}$ & 85.04$_{\pm 0.016}$ & 85.52$_{\pm 0.009}$ & 84.97$_{\pm 0.003}$ & 84.99$_{\pm 0.004}$ & \underline{88.13}$_{\pm 0.007}$ & \textbf{88.14}$_{\pm 0.007}$ & <0.05 & <0.05 \\  
ivy      & 24.20$_{\pm 0.005}$ & 17.65$_{\pm 0.004}$ & 17.69$_{\pm 0.002}$ & {26.99}$_{\pm 0.013}$ & {28.02}$_{\pm 0.013}$ & \textbf{33.22}$_{\pm 0.013}$ & \underline{30.47}$_{\pm 0.006}$ & 25.06$_{\pm 0.004}$ & 25.25$_{\pm 0.004}$ & <0.05 & <0.05 \\  
prop6    & 17.28$_{\pm 0.010}$ & 14.37$_{\pm 0.001}$ & 14.48$_{\pm 0.002}$ & {19.90}$_{\pm 0.009}$ & {20.19}$_{\pm 0.016}$ & \underline{21.48}$_{\pm 0.005}$ & \textbf{22.01}$_{\pm 0.007}$ & 19.19$_{\pm 0.004}$ & 19.26$_{\pm 0.002}$ & <0.05 & <0.05 \\  
redaktor & \textbf{21.61}$_{\pm 0.012}$ & 11.79$_{\pm 0.006}$ & 12.55$_{\pm 0.009}$ & 18.92$_{\pm 0.011}$ & \underline{20.69}$_{\pm 0.027}$ & 14.06$_{\pm 0.007}$ & 13.80$_{\pm 0.006}$ & 19.71$_{\pm 0.015}$ & 20.20$_{\pm 0.012}$ & 0.306 & <0.05 \\ 

tomcat   & 19.07$_{\pm 0.007}$ & 12.10$_{\pm 0.001}$ & 11.97$_{\pm 0.002}$ & 22.30$_{\pm 0.006}$ & 21.30$_{\pm 0.009}$ & 19.46$_{\pm 0.006}$ & 17.38$_{\pm 0.004}$ & \underline{23.14}$_{\pm 0.004}$ & \textbf{23.15}$_{\pm 0.005}$ & <0.05 & <0.05 \\ \hline  

\rowcolor{pink!30} Avg. \& W/T/L           & 47.26$_{\pm 0.012}$ & 37.55$_{\pm 0.005}$ & 37.52$_{\pm 0.006}$ & 48.96$_{\pm 0.018}$ & 48.96$_{\pm 0.015}$ &\textbf{49.25}$_{\pm 0.013}$ &46.81$_{\pm 0.0011}$ &   {49.00}$_{\pm 0.009}$ & \underline{49.02}$_{\pm 0.008}$ &5/3/5 &7/1/5\\ 
\hline

 ar3& 25.90$_{\pm 0.025}$ & 26.62$_{\pm 0.013}$ & 27.37$_{\pm 0.017}$ & 26.83$_{\pm 0.021}$ & 26.53$_{\pm 0.026}$ & 25.40$_{\pm 0.006}$ & 25.87$_{\pm 0.005}$ & \underline{28.00}$_{\pm 0.002}$ & \textbf{28.34}$_{\pm 0.003}$ & <0.05 & <0.05 \\ 
ar4 & 37.54$_{\pm 0.018}$ & 34.38$_{\pm 0.052}$ & 34.69$_{\pm 0.050}$ & 34.71$_{\pm 0.043}$ & 35.30$_{\pm 0.038}$ & \underline{40.07}$_{\pm 0.011}$ & \textbf{40.44}$_{\pm 0.010}$ & 38.89$_{\pm 0.003}$ & 39.12$_{\pm 0.035}$ & <0.05 & <0.05 \\
ar5 & 60.68$_{\pm 0.023}$ & \underline{67.84}$_{\pm 0.079}$ & \textbf{68.29}$_{\pm 0.079}$ & 65.47$_{\pm 0.098}$ & 64.92$_{\pm 0.112}$ & 53.19$_{\pm 0.055}$ & 53.85$_{\pm 0.052}$ & 60.12$_{\pm 0.055}$ & 60.66$_{\pm 0.060}$ & <0.05 & <0.05 \\ \
ar6 & 22.71$_{\pm 0.018}$ & 19.91$_{\pm 0.024}$ & 20.06$_{\pm 0.024}$ & 22.65$_{\pm 0.037}$ & 23.33$_{\pm 0.028}$ & 21.84$_{\pm 0.034}$ & 21.98$_{\pm 0.033}$ & \underline{27.93}$_{\pm 0.012}$ & \textbf{28.53}$_{\pm 0.020}$ & <0.05 & <0.05 \\\hline
\rowcolor{pink!30}{Avg. \& W/T/L} & 36.71$_{\pm 0.021}$ & 37.19$_{\pm 0.042}$ & 37.60$_{\pm 0.043}$ & 37.42$_{\pm 0.050}$ & 37.52$_{\pm 0.051}$ & 35.13$_{\pm 0.027}$ & 35.53$_{\pm 0.025}$ & \underline{38.74}$_{\pm 0.018}$ & \textbf{39.17}$_{\pm 0.030}$ & 3/0/1 & 3/0/1 \\ 

\toprule
\end{tabular}
}
\end{table*}

\begin{table*}[!h]
\caption{Comparison of FedDP and baseline methods in terms of Recall value.\label{tab:table2}}
\centering
\resizebox{\columnwidth}{!}{
\begin{tabular}{cccccccc>{\columncolor{blue!10} }c>{\columncolor{blue!10} }ccc}
% \begin{tabular*}{\textwidth}{@{\extracolsep\fill}c|c|ccccc|cc@{}}
\bottomrule
\rowcolor{white}
  \multirow{3}{*}{Project}& Recall &  &  &  &  &  &  & &  &$p$-value&\\ 
\cmidrule(r){2-10} \cmidrule(r){11-12}
 & LR & OpenFLR & OpenFLR & FLR & FLR &Almity &Almity & \cellcolor{white}{FedDP} & \cellcolor{white}{FedDP} & vs. FLR &vs. Almity\\  

 & Centralized &FedAvg & FedProx & FedAvg & FedProx & FedAvg & FedProx& \cellcolor{white}FedAvg & \cellcolor{white}FedProx &FedProx &FedProx\\  
\hline

ant & 73.43$_{\pm 0.019}$ & \underline{86.17}$_{\pm 0.037}$ &  \textbf{88.24}$_{\pm 0.013}$ & 71.57$_{\pm 0.015}$ & 71.20$_{\pm 0.014}$ & 66.99$_{\pm 0.021}$ & 73.37$_{\pm 0.015}$ & 74.88$_{\pm 0.009}$ & 74.88$_{\pm 0.007}$ & <0.05 & <0.05 \\

jedit & 72.91$_{\pm 0.042}$ & \textbf{84.25}$_{\pm 0.008}$ & \underline{81.82}$_{\pm 0.033}$ & 70.30$_{\pm 0.021}$ & 71.57$_{\pm 0.020}$ & 71.90$_{\pm 0.011}$ & 74.18$_{\pm 0.011}$ & 71.49$_{\pm 0.005}$ & 70.99$_{\pm 0.005}$ & <0.05 & <0.05 \\

lucene & 47.98$_{\pm 0.046}$ & \textbf{58.66}$_{\pm 0.052}$ & \underline{58.50}$_{\pm 0.052}$ & 40.04$_{\pm 0.034}$ & 39.88$_{\pm 0.025}$ & 38.47$_{\pm 0.022}$ & 42.56$_{\pm 0.011}$ & 43.78$_{\pm 0.018}$ & 44.71$_{\pm 0.023}$ & <0.05 & <0.05 \\

xerces & \underline{51.08}$_{\pm 0.037}$ & \textbf{52.09}$_{\pm 0.027}$ & 50.00$_{\pm 0.031}$ & 39.69$_{\pm 0.053}$ & 38.28$_{\pm 0.022}$ & 45.85$_{\pm 0.013}$ & 47.69$_{\pm 0.011}$ & 45.91$_{\pm 0.005}$ & 45.97$_{\pm 0.010}$ & <0.05 & <0.05 \\

velocity & 37.18$_{\pm 0.013}$ & \textbf{54.03}$_{\pm 0.015}$ & \underline{50.59}$_{\pm 0.037}$ & 33.85$_{\pm 0.018}$ & 33.59$_{\pm 0.012}$ & 30.77$_{\pm 0.024}$ & 35.13$_{\pm 0.025}$ & 41.90$_{\pm 0.014}$ & 42.03$_{\pm 0.008}$ & <0.05 & <0.05 \\

xalan & 48.13$_{\pm 0.008}$ & \textbf{67.11}$_{\pm 0.060}$ & \underline{67.05}$_{\pm 0.080}$ & 47.36$_{\pm 0.027}$ & 45.72$_{\pm 0.023}$ & 47.93$_{\pm 0.014}$ & 50.61$_{\pm 0.009}$ & 49.27$_{\pm 0.009}$ & 49.01$_{\pm 0.007}$ & <0.05 & <0.05 \\

synapse & 66.74$_{\pm 0.021}$ & \textbf{72.09}$_{\pm 0.018}$ & \underline{71.51}$_{\pm 0.020}$ & 55.77$_{\pm 0.010}$ & 57.12$_{\pm 0.007}$ & 50.70$_{\pm 0.016}$ & 53.26$_{\pm 0.015}$ & 62.91$_{\pm 0.020}$ & 62.21$_{\pm 0.014}$ & <0.05 & <0.05 \\

log4j & 74.07$_{\pm 0.026}$ & \underline{86.00}$_{\pm 0.009}$ & \textbf{87.04}$_{\pm 0.017}$ & 60.30$_{\pm 0.090}$ & 61.33$_{\pm 0.037}$ & 55.56$_{\pm 0.003}$ & 57.78$_{\pm 0.020}$ & 57.04$_{\pm 0.017}$ & 57.33$_{\pm 0.024}$ & <0.05 & 0.524 \\
poi & 52.24$_{\pm 0.025}$ &\underline{69.00}$_{\pm 0.035}$ & \textbf{76.71}$_{\pm 0.041}$ & 36.78$_{\pm 0.018}$ & 36.14$_{\pm 0.016}$ & 41.85$_{\pm 0.022}$ & 48.75$_{\pm 0.009}$ & 50.18$_{\pm 0.014}$ & 51.54$_{\pm 0.010}$ & <0.05 & <0.05 \\

ivy & \textbf{79.50}$_{\pm 0.021}$ & 74.90$_{\pm 0.034}$ & \underline{79.30}$_{\pm 0.033}$ & 74.30$_{\pm 0.018}$ & 76.55$_{\pm 0.020}$ & 73.50$_{\pm 0.029}$ & 76.00$_{\pm 0.014}$ & 77.10$_{\pm 0.009}$ & 77.50$_{\pm 0.010}$ & <0.05 & <0.05 \\

prop6 & 58.79$_{\pm 0.013}$ & 59.55$_{\pm 0.013}$ & 57.21$_{\pm 0.011}$ & 48.91$_{\pm 0.039}$ & 49.82$_{\pm 0.034}$ & 40.91$_{\pm 0.036}$ & 47.88$_{\pm 0.038}$ & \textbf{61.39}$_{\pm 0.020}$ & \underline{60.45}$_{\pm 0.008}$ & <0.05 & <0.05 \\

redaktor & \textbf{71.11}$_{\pm 0.031}$ & 29.11$_{\pm 0.030}$ & 32.74$_{\pm 0.021}$ & 47.48$_{\pm 0.050}$ & 53.93$_{\pm 0.081}$ & 48.15$_{\pm 0.037}$ & 50.37$_{\pm 0.020}$ & 66.15$_{\pm 0.059}$ & \underline{69.04}$_{\pm 0.052}$ & <0.05 & <0.05 \\

tomcat & 84.94$_{\pm 0.020}$ & \underline{89.69}$_{\pm 0.016}$ & \textbf{91.40}$_{\pm 0.006}$ & 79.87$_{\pm 0.050}$ & 78.99$_{\pm 0.089}$ & 79.74$_{\pm 0.012}$ & 82.60$_{\pm 0.007}$ & 72.78$_{\pm 0.018}$ & 72.65$_{\pm 0.017}$ & <0.05 & <0.05 \\
\hline
\rowcolor{pink!30}  Avg. \& W/T/L & 62.93$_{\pm 0.025}$ & \underline{67.90}$_{\pm 0.027}$ & \textbf{68.62}$_{\pm 0.030}$ & 54.32$_{\pm 0.034}$ & 54.93$_{\pm 0.031}$ & 53.25$_{\pm 0.020}$ & 56.94$_{\pm 0.016}$ & 59.60$_{\pm 0.017}$ & 59.87$_{\pm 0.015}$ & 10/0/3 & 8/1/4 \\
  
\hline  
ar3 & 77.50$_{\pm 0.056}$ & 72.12$_{\pm 0.056}$ & 72.50$_{\pm 0.056}$ & 72.93$_{\pm 0.061}$ & 72.50$_{\pm 0.056}$ & 75.30$_{\pm 0.060}$ & 75.00$_{\pm 0.064}$ & \textbf{80.15}$_{\pm 0.075}$ & \underline{80.00}$_{\pm 0.068}$ & <0.05 & <0.05 \\   
ar4 & \textbf{81.00}$_{\pm 0.022}$ & 45.23$_{\pm 0.050}$ & 45.70$_{\pm 0.043}$ & 52.68$_{\pm 0.068}$ & 52.20$_{\pm 0.074}$ & \underline{74.20}$_{\pm 0.019}$ & 74.00$_{\pm 0.022}$ & 48.87$_{\pm 0.051}$ & 49.30$_{\pm 0.045}$ & <0.05 & <0.05 \\   
ar5 & \textbf{92.50}$_{\pm 0.068}$ & 86.90$_{\pm 0.084}$ & 87.50$_{\pm 0.088}$ & 86.05$_{\pm 0.011}$ & 87.25$_{\pm 0.006}$ & 86.70$_{\pm 0.039}$ & 87.50$_{\pm 0.037}$ & \underline{91.04}$_{\pm 0.088}$ & 90.50$_{\pm 0.096}$ & <0.05 & <0.05 \\   
ar6 & \textbf{60.00}$_{\pm 0.067}$ & 39.68$_{\pm 0.065}$ & 40.00$_{\pm 0.073}$ & 48.16$_{\pm 0.068}$ & 49.33$_{\pm 0.060}$ & 52.84$_{\pm 0.052}$ & \underline{53.32}$_{\pm 0.055}$ & 52.96$_{\pm 0.030}$ & 53.33$_{\pm 0.023}$ & <0.05 & 0.967 \\ \hline  
\rowcolor{pink!30}  Avg. \& W/T/L & \textbf{77.75}$_{\pm 0.053}$ & 60.98$_{\pm 0.064}$ & 61.43$_{\pm 0.065}$ & 64.96$_{\pm 0.052}$ & 65.32$_{\pm 0.049}$ & 72.26$_{\pm 0.043}$ & \underline{72.46}$_{\pm 0.045}$ & 68.26$_{\pm 0.061}$ & 68.28$_{\pm 0.058}$ & 3/0/1 & 2/1/1 \\ 

\toprule
\end{tabular}}
\end{table*}

\begin{table*}[!h]
\caption{Comparison of FedDP and baseline methods in terms of F1 value.\label{tab:table3}}
\centering
\resizebox{\columnwidth}{!}{
\begin{tabular}{cccccccc>{\columncolor{blue!10} }c>{\columncolor{blue!10} }ccc}
% \begin{tabular*}{\textwidth}{@{\extracolsep\fill}c|c|ccccc|cc@{}}
\bottomrule
\rowcolor{white}
  \multirow{3}{*}{Project}& F1 &  &  &  &  &  &  & &  &$p$-value&\\ 
\cmidrule(r){2-10} \cmidrule(r){11-12}
 & LR & OpenFLR & OpenFLR & FLR & FLR &Almity &Almity & \cellcolor{white}{FedDP} & \cellcolor{white}{FedDP} & vs. FLR &vs. Almity\\  

 & Centralized &FedAvg & FedProx & FedAvg & FedProx & FedAvg & FedProx& \cellcolor{white}FedAvg & \cellcolor{white}FedProx &FedProx &FedProx\\  
\hline

ant & 54.63$_{\pm 0.005}$ & 47.93$_{\pm 0.005}$ & 48.62$_{\pm 0.005}$ & \underline{57.46}$_{\pm 0.005}$ & \textbf{58.33}$_{\pm 0.009}$ & 57.61$_{\pm 0.008}$ & 57.13$_{\pm 0.003}$ & 56.39$_{\pm 0.006}$ & 56.57$_{\pm 0.008}$ & <0.05 & <0.05 \\

jedit & 55.41$_{\pm 0.013}$ & 50.13$_{\pm 0.002}$ & 49.58$_{\pm 0.007}$ & 57.77$_{\pm 0.015}$ & 57.00$_{\pm 0.009}$ & \textbf{62.43}$_{\pm 0.007}$ & \underline{59.45}$_{\pm 0.010}$ & 58.39$_{\pm 0.007}$ & 58.39$_{\pm 0.004}$ & <0.05 & <0.05 \\

lucene & 58.50$_{\pm 0.034}$ & \textbf{62.94}$_{\pm 0.036}$ & \underline{62.78}$_{\pm 0.036}$ & 52.54$_{\pm 0.025}$ & 52.50$_{\pm 0.023}$ & 50.64$_{\pm 0.017}$ & 53.52$_{\pm 0.009}$ & 55.94$_{\pm 0.015}$ & 56.50$_{\pm 0.017}$ & <0.05 & <0.05 \\

xerces & \textbf{45.88}$_{\pm 0.019}$ & 28.02$_{\pm 0.010}$ & 27.72$_{\pm 0.012}$ & 39.02$_{\pm 0.044}$ & 38.02$_{\pm 0.014}$ & 40.06$_{\pm 0.011}$ & 38.71$_{\pm 0.011}$ & 40.43$_{\pm 0.005}$ & \underline{40.52}$_{\pm 0.006}$ & <0.05 & <0.05 \\

velocity & 43.80$_{\pm 0.008}$ & \textbf{49.82}$_{\pm 0.007}$ & \underline{48.46}$_{\pm 0.020}$ & 42.21$_{\pm 0.014}$ & 41.85$_{\pm 0.014}$ & 39.18$_{\pm 0.023}$ & 41.93$_{\pm 0.021}$ & 47.70$_{\pm 0.010}$ & 47.62$_{\pm 0.006}$ & <0.05 & <0.05 \\

xalan & 51.48$_{\pm 0.004}$ & \textbf{64.33}$_{\pm 0.031}$ & \underline{64.02}$_{\pm 0.048}$ & 51.67$_{\pm 0.017}$ & 50.21$_{\pm 0.015}$ & 56.60$_{\pm 0.007}$ & 57.30$_{\pm 0.007}$ & 52.94$_{\pm 0.006}$ & 52.72$_{\pm 0.004}$ & <0.05 & <0.05 \\

synapse & \textbf{64.43}$_{\pm 0.011}$ & 55.54$_{\pm 0.006}$ & 55.15$_{\pm 0.006}$ & 60.02$_{\pm 0.010}$ & 60.91$_{\pm 0.006}$ & 58.75$_{\pm 0.012}$ & 59.18$_{\pm 0.009}$ & 61.32$_{\pm 0.003}$ & \underline{61.36}$_{\pm 0.004}$ & <0.05 & <0.05 \\

log4j & \textbf{72.46}$_{\pm 0.004}$ & 56.42$_{\pm 0.007}$ & 55.23$_{\pm 0.010}$ & 65.97$_{\pm 0.050}$ & 66.03$_{\pm 0.021}$ & 56.90$_{\pm 0.025}$ & 55.70$_{\pm 0.007}$ &\underline{66.66}$_{\pm 0.011}$ & 66.56$_{\pm 0.014}$ & 0.448 & <0.05 \\

poi & 64.64$_{\pm 0.020}$ & \underline{70.59}$_{\pm 0.025}$ & \textbf{75.34}$_{\pm 0.024}$ & 51.31$_{\pm 0.016}$ & 50.78$_{\pm 0.017}$ & 56.05$_{\pm 0.020}$ & 61.96$_{\pm 0.007}$ & 63.85$_{\pm 0.010}$ & 64.99$_{\pm 0.009}$ & <0.05 & <0.05 \\

ivy & 37.10$_{\pm 0.008}$ & 28.44$_{\pm 0.005}$ & 28.90$_{\pm 0.005}$ & 39.59$_{\pm 0.013}$ & 40.99$_{\pm 0.017}$ & \textbf{45.74}$_{\pm 0.014}$ & \underline{43.50}$_{\pm 0.007}$ & 37.82$_{\pm 0.004}$ & 38.08$_{\pm 0.005}$ & <0.05 & <0.05 \\

prop6 & 26.70$_{\pm 0.013}$ & 23.09$_{\pm 0.001}$ & 22.97$_{\pm 0.002}$ & 28.24$_{\pm 0.014}$ & 28.71$_{\pm 0.022}$ & 28.14$_{\pm 0.012}$ & \textbf{30.14}$_{\pm 0.014}$ & \underline{29.22}$_{\pm 0.005}$ & 29.20$_{\pm 0.002}$ & <0.05 & <0.05 \\

redaktor & \textbf{33.11}$_{\pm 0.013}$ & 16.27$_{\pm 0.009}$ & 17.73$_{\pm 0.012}$ & 26.99$_{\pm 0.017}$ & 29.88$_{\pm 0.040}$ & 21.76$_{\pm 0.012}$ & 21.66$_{\pm 0.009}$ & 30.36$_{\pm 0.023}$ & \underline{31.24}$_{\pm 0.020}$ & <0.05 & <0.05 \\

tomcat & 31.14$_{\pm 0.008}$ & 21.30$_{\pm 0.002}$ & 21.16$_{\pm 0.002}$ & 34.80$_{\pm 0.006}$ & 33.53$_{\pm 0.011}$ & 31.28$_{\pm 0.007}$ & 28.72$_{\pm 0.006}$ & \underline{35.08}$_{\pm 0.006}$ & \textbf{35.09}$_{\pm 0.007}$ & <0.05 & <0.05 \\
\hline
 \rowcolor{pink!30}  Avg. \& W/T/L & \textbf{49.18}$_{\pm 0.012}$ & 44.22$_{\pm 0.011}$ & 44.44$_{\pm 0.015}$ & 46.74$_{\pm 0.019}$ & 46.83$_{\pm 0.017}$ & 46.55$_{\pm 0.013}$ & 46.84$_{\pm 0.009}$ & 48.93$_{\pm 0.009}$ & \underline{49.14}$_{\pm 0.008}$ & 10/1/2 &  8/0/5 \\
\hline
ar3 & 38.69$_{\pm 0.025}$ & 40.01$_{\pm 0.015}$ & 39.66$_{\pm 0.020}$ & 38.66$_{\pm 0.025}$ & 38.83$_{\pm 0.035}$ & 37.99$_{\pm 0.004}$ & 38.47$_{\pm 0.005}$ & \underline{41.47}$_{\pm 0.004}$ & \textbf{41.80}$_{\pm 0.006}$ & <0.05 & <0.05 \\   
ar4 & 51.29$_{\pm 0.020}$ & 38.93$_{\pm 0.051}$ & 39.38$_{\pm 0.047}$ & 41.69$_{\pm 0.063}$ & 42.09$_{\pm 0.050}$ & \underline{52.03}$_{\pm 0.010}$ & \textbf{52.29}$_{\pm 0.013}$ & 43.25$_{\pm 0.041}$ & 43.56$_{\pm 0.035}$ & <0.05 & <0.05 \\   
ar5 & 73.21$_{\pm 0.031}$ & \underline{75.87}$_{\pm 0.068}$ & \textbf{76.40}$_{\pm 0.067}$ & 73.72$_{\pm 0.077}$ & 74.02$_{\pm 0.073}$ & 65.91$_{\pm 0.045}$ & 66.67$_{\pm 0.041}$ & 72.01$_{\pm 0.053}$ & 72.28$_{\pm 0.049}$ & <0.05 & <0.05 \\   
ar6 & 32.93$_{\pm 0.028}$ & 26.04$_{\pm 0.021}$ & 26.65$_{\pm 0.021}$ & 32.04$_{\pm 0.031}$ & 31.64$_{\pm 0.036}$ & 30.91$_{\pm 0.024}$ & 31.13$_{\pm 0.023}$ & \underline{36.60}$_{\pm 0.012}$ & \textbf{37.14}$_{\pm 0.017}$ & <0.05 & <0.05 \\   
\hline
\rowcolor{pink!30}  
Avg. \& W/T/L & \textbf{49.03}$_{\pm 0.026}$ & 45.21$_{\pm 0.039}$ & 45.52$_{\pm 0.039}$ & 46.53$_{\pm 0.049}$ & 46.64$_{\pm 0.048}$ & 46.71$_{\pm 0.021}$ & 47.14$_{\pm 0.021}$ & 48.33$_{\pm 0.028}$ & \underline{48.69}$_{\pm 0.027}$ & 3/0/1 & 3/0/1 \\   
\hline

\toprule
\end{tabular}}
\end{table*}

\begin{table*}[!h]
\caption{Comparison of FedDP and baseline methods in terms of AUC value.\label{tab:table4}}
\centering
\resizebox{\columnwidth}{!}{
\begin{tabular}{cccccccc>{\columncolor{blue!10} }c>{\columncolor{blue!10} }ccc}
% \begin{tabular*}{\textwidth}{@{\extracolsep\fill}c|c|ccccc|cc@{}}
\bottomrule
\rowcolor{white}
  \multirow{3}{*}{Project}& AUC &  &  &  &  &  &  & &  &$p$-value&\\ 
\cmidrule(r){2-10} \cmidrule(r){11-12}
 & LR & OpenFLR & OpenFLR & FLR & FLR &Almity &Almity & \cellcolor{white}{FedDP} & \cellcolor{white}{FedDP} & vs. FLR &vs. Almity\\  

 & Centralized &FedAvg & FedProx & FedAvg & FedProx & FedAvg& FedProx& \cellcolor{white}FedAvg & \cellcolor{white}FedProx &FedProx &FedProx\\  
\hline
ant & 73.04$_{\pm 0.003}$ & 68.30$_{\pm 0.005}$ & 69.07$_{\pm 0.006}$ & 74.65$_{\pm 0.003}$ & \textbf{75.14}$_{\pm 0.007}$ & 74.10$_{\pm 0.007}$ & 74.72$_{\pm 0.002}$ & \underline{74.42}$_{\pm 0.003}$ & 74.54$_{\pm 0.004}$ & <0.05 & 0.436 \\
jedit & 71.18$_{\pm 0.013}$ & 66.45$_{\pm 0.002}$ & 65.99$_{\pm 0.006}$ & 72.71$_{\pm 0.011}$ & 72.28$_{\pm 0.007}$ & \textbf{76.04}$_{\pm 0.004}$ & \underline{74.30}$_{\pm 0.007}$ & {73.28}$_{\pm 0.005}$ & 73.22$_{\pm 0.003}$ & <0.05 & <0.05 \\
lucene & 62.23$_{\pm 0.013}$ & 60.77$_{\pm 0.008}$ & 60.54$_{\pm 0.007}$ & 60.81$_{\pm 0.011}$ & 60.99$_{\pm 0.012}$ & 59.24$_{\pm 0.004}$ & 59.00$_{\pm 0.004}$ & \textbf{62.49}$_{\pm 0.008}$ & \underline{62.37}$_{\pm 0.006}$ & <0.05 & <0.05 \\
xerces & \textbf{68.91}$_{\pm 0.015}$ & 56.17$_{\pm 0.010}$ & 55.90$_{\pm 0.010}$ & 63.93$_{\pm 0.028}$ & 63.28$_{\pm 0.009}$ & 65.21$_{\pm 0.007}$ & 64.67$_{\pm 0.007}$ & \underline{65.41}$_{\pm 0.003}$ & 65.47$_{\pm 0.004}$ & <0.05 & <0.05 \\
velocity & 60.18$_{\pm 0.006}$ & 61.14$_{\pm 0.004}$ & 60.71$_{\pm 0.007}$ & 60.08$_{\pm 0.006}$ & 59.93$_{\pm 0.008}$ & 58.63$_{\pm 0.011}$ & 59.22$_{\pm 0.010}$ & \textbf{62.27}$_{\pm 0.005}$ & \underline{62.13}$_{\pm 0.004}$ & <0.05 & <0.05 \\
xalan & 57.23$_{\pm 0.006}$ & \underline{66.53}$_{\pm 0.013}$ & \textbf{66.73}$_{\pm 0.021}$ & 58.11$_{\pm 0.006}$ & 57.09$_{\pm 0.009}$ & 64.68$_{\pm 0.003}$ & 64.02$_{\pm 0.008}$ & 58.72$_{\pm 0.003}$ & 58.58$_{\pm 0.002}$ & <0.05 & <0.05 \\
synapse & \textbf{73.14}$_{\pm 0.008}$ & 64.02$_{\pm 0.004}$ & 63.66$_{\pm 0.004}$ & 70.27$_{\pm 0.007}$ & \underline{70.86}$_{\pm 0.004}$ & 69.82$_{\pm 0.007}$ & 69.86$_{\pm 0.007}$ & 70.77$_{\pm 0.002}$ & 70.85$_{\pm 0.003}$ & 0.911 & <0.05 \\
log4j & \textbf{79.72}$_{\pm 0.004}$ & 64.23$_{\pm 0.009}$ & 62.45$_{\pm 0.015}$ & 74.72$_{\pm 0.032}$ & 74.93$_{\pm 0.015}$ & 68.13$_{\pm 0.020}$ & 66.92$_{\pm 0.004}$ & \underline{75.14}$_{\pm 0.007}$ & 75.11$_{\pm 0.009}$ & 0.702 & <0.05 \\
poi & 67.98$_{\pm 0.010}$ & 64.60$_{\pm 0.007}$ & 66.19$_{\pm 0.011}$ & 62.72$_{\pm 0.007}$ & 62.71$_{\pm 0.009}$ & 64.47$_{\pm 0.008}$ & 66.86$_{\pm 0.003}$ & \underline{69.17}$_{\pm 0.005}$ & \textbf{69.70}$_{\pm 0.006}$ & <0.05 & <0.05 \\
ivy & 73.79$_{\pm 0.010}$ & 65.05$_{\pm 0.006}$ & 66.01$_{\pm 0.009}$ & 74.25$_{\pm 0.010}$ & 75.62$_{\pm 0.015}$ & \textbf{77.26}$_{\pm 0.012}$ & \underline{76.88}$_{\pm 0.006}$ & 73.76$_{\pm 0.003}$ & 74.02$_{\pm 0.005}$ & <0.05 & <0.05 \\
prop6 & 63.72$_{\pm 0.013}$ & 60.04$_{\pm 0.001}$ & 59.80$_{\pm 0.003}$ & 63.51$_{\pm 0.015}$ & 63.97$_{\pm 0.020}$ & 62.15$_{\pm 0.013}$ & 64.53$_{\pm 0.015}$ & \textbf{66.31}$_{\pm 0.006}$ & \underline{66.13}$_{\pm 0.002}$ & <0.05 & <0.05 \\
redaktor & \textbf{62.07}$_{\pm 0.015}$ & 46.15$_{\pm 0.013}$ & 46.22$_{\pm 0.010}$ & 55.44$_{\pm 0.018}$ & 58.24$_{\pm 0.045}$ & 47.43$_{\pm 0.014}$ & 46.66$_{\pm 0.012}$ & 58.70$_{\pm 0.030}$ & \underline{59.83}$_{\pm 0.026}$ & <0.05 & <0.05 \\
tomcat & 74.67$_{\pm 0.004}$ & 62.58$_{\pm 0.002}$ & 62.48$_{\pm 0.004}$ & \textbf{76.05}$_{\pm 0.015}$ & \underline{75.10}$_{\pm 0.027}$ & 73.58$_{\pm 0.003}$ & 71.94$_{\pm 0.007}$ & 74.43$_{\pm 0.008}$ & 74.41$_{\pm 0.007}$ & <0.05 & <0.05 \\
\hline
 \rowcolor{pink!30}  Avg. \& W/T/L & \textbf{68.30}$_{\pm 0.009}$ & 62.00$_{\pm 0.006}$ & 61.98$_{\pm 0.009}$ & 66.71$_{\pm 0.013}$ & 66.93$_{\pm 0.014}$ & 66.21$_{\pm 0.009}$ & 66.12$_{\pm 0.007}$ & 68.07$_{\pm 0.007}$ & \underline{68.18}$_{\pm 0.006}$ & 8/2/3 & 9/1/3 \\ \hline

ar3 & 72.39$_{\pm 0.016}$ & 72.09$_{\pm 0.020}$ & 72.18$_{\pm 0.020}$ & 71.28$_{\pm 0.036}$ & 71.60$_{\pm 0.034}$ & 72.19$_{\pm 0.023}$ & 71.86$_{\pm 0.024}$ & \underline{74.98}$_{\pm 0.016}$ & \textbf{75.27}$_{\pm 0.019}$ & <0.05 & <0.05 \\   
ar4 & \textbf{74.98}$_{\pm 0.017}$ & 63.13$_{\pm 0.038}$ & 62.83$_{\pm 0.035}$ & 64.89$_{\pm 0.038}$ & 65.13$_{\pm 0.039}$ & 73.92$_{\pm 0.015}$ & \underline{74.47}$_{\pm 0.012}$ & 66.22$_{\pm 0.020}$ & 65.80$_{\pm 0.024}$ & 0.283 & <0.05 \\   
ar5 & \underline{87.68}$_{\pm 0.032}$ & 86.60$_{\pm 0.048}$ & \textbf{87.75}$_{\pm 0.048}$ & 87.05$_{\pm 0.028}$ & 86.45$_{\pm 0.033}$ & 82.50$_{\pm 0.014}$ & 83.04$_{\pm 0.016}$ & 86.31$_{\pm 0.038}$ & 86.68$_{\pm 0.043}$ & 0.747 & <0.05 \\   
ar6 & 62.21$_{\pm 0.031}$ & 55.86$_{\pm 0.020}$ & 55.91$_{\pm 0.020}$ & 59.95$_{\pm 0.050}$ & 60.49$_{\pm 0.031}$ & 60.46$_{\pm 0.005}$ & 60.16$_{\pm 0.003}$ & \underline{64.70}$_{\pm 0.017}$ & \textbf{64.95}$_{\pm 0.012}$ & <0.05 & <0.05 \\
\hline
\rowcolor{pink!30}  
Avg. \& W/T/L & \textbf{74.31}$_{\pm 0.024}$ & 69.42$_{\pm 0.032}$ & 69.67$_{\pm 0.031}$ & 70.79$_{\pm 0.038}$ & 70.92$_{\pm 0.034}$ & 72.27$_{\pm 0.014}$ & 72.38$_{\pm 0.014}$ & 73.05$_{\pm 0.023}$ & \underline{73.17}$_{\pm 0.025}$ & 2/2/0 & 3/0/1 \\   
\toprule
\end{tabular}}
\end{table*}
We evaluate the predictive performance of FedDP and baseline methods with Precision, Recall, F1, and AUC four metrics on two datasets. Additionally, given that our method significantly outperforms OpenFLR, we conduct significance tests with FLR and Almity. The results are shown in Tables \ref{tab:table1} to \ref{tab:table4}. The column `Project' corresponds to the specific test project, indicating that the latest version of data from this project is used as the test set. We represent the mean and standard deviation of five independent experiments conducted on each test set. We use bold type to highlight the optimal and underlining to highlight the suboptimal results.

Centralized training, as the upper bound for FL performance, achieves the best results in the comprehensive evaluation metrics of F1 and AUC in all datasets. This indicates that when data can be processed centrally, the performance of the model is often superior because centralized training has access to all distributed data, allowing the model to learn and capture features of the data more comprehensively. It is worth noting that FedDP achieves suboptimal results in terms of F1 and AUC, closely resembling the performance of centralized training. This underscores the ability of FedDP to collaboratively train an effective model in a distributed manner while safeguarding local data privacy.

On the Promise dataset, FedDP outperforms FL-based methods by an average of 3.10\% in F1 and 3.13\% in AUC. Similarly, on the Softlab dataset, FedDP improves by an average of 2.22\% in F1 and 2.21\% in AUC. Although OpenFLR obtains the best performance in terms of Recall on Promise, it sacrifices the Precision, resulting in more false positives. With no accident, OpenFLR demonstrates the worst performance among all evaluated methods. As analyzed in Section \ref{openflr_analysis}, consolidating the global model by training with open-source project data suffers the variance in the optimization direction and the disadvantages of data heterogeneity outweigh the advantages of additional knowledge from the open-source project data. In contrast, FedDP effectively mitigates this risk by leveraging knowledge from open-source project data through knowledge distillation. Notably, Almity introduces improvements to FedAvg for dealing with uneven data distributions, but it fundamentally belongs to the primary parameter aggregation approach, performing like FLR and worse than FedDP.  This underscores that our knowledge distillation framework is significantly more effective in mitigating the challenges posed by Non-IID data than parameter aggregation methods.
% different algorithms
In addition, FedDP demonstrates excellent performance across different FL algorithms (whether FedAvg or FedProx) and datasets with differing Non-IID degrees and programming languages. This underscores FedDP's robustness. Besides, our method exhibits smaller standard deviations and achieves a stable performance.

% conclusion (revised by lyc) 
In conclusion, FedDP outperforms other FL-based CPDP methods in terms of the comprehensive evaluation metrics F1 and AUC, and its performance is closer to centralized training. Furthermore, FedDP exhibits minimal performance fluctuations with different test projects and stays robust to varying FL algorithms and datasets.

\subsection{RQ2: How does the communication efficiency of FedDP compare to that of other FL-based CPDP methods? }
Unlike centralized environments, in FL, the communication overhead is an expensive expense and often serves as a key indicator for method evaluation. In this RQ, we compare the communication efficiency of FL-based CPDP methods by evaluating the number of communication rounds required to stably reach the target F1, as shown in Table \ref{tab:table5}. For each test project, we set two F1 thresholds for evaluation based on the average performance of all methods. We bold the optimal results and record the times of each method achieving the best performance in the last row of the table, denoted as `Total'.

The experimental results demonstrate that FedDP exhibits superior communication efficiency compared to other methods, requiring fewer communication rounds to achieve target performance. Furthermore, across all 13 test items, when increasing the target performance threshold, FedDP requires minimal additional communication rounds, typically within two rounds to achieve the target performance. In contrast, in most cases, FLR typically requires more than two additional rounds, while OpenFLR and Almity often fail to stably reach the target performance even after 50 rounds of communication. Consequently, OpenFLR and FLR require more significant effort and communication costs than FedDP to attain the same target performances.
In conclusion, FedDP demonstrates significant advantages in communication efficiency compared to other FL-based CPDP methods, rapidly achieving target performance with fewer communication rounds.

\begin{table*}[t]
\caption{Evaluation of FL-based CPDP methods, in terms of the communication rounds to reach the target F1 value.\label{tab:table5}}
\centering

\begin{tabular}{ccccccccccc}
\bottomrule
\multirow{2}{*}{ Project} & \multirow{2}{*}{ F1} &  OpenFLR &  OpenFLR &  FLR &  FLR &Almity &Almity & FedDP &  FedDP \\  
 & &  FedAvg &  FedProx &  FedAvg &  FedProx &  FedAvg &  FedProx &  FedAvg &  FedProx \\  
\hline

\multirow{2}{*}{ant} & 52.5\% & \textgreater{}50 & \textgreater{}50 & 10.5 & 7.8 & \textbf{2.0} & 2.4 & 2.2 & \textbf{2.0} \\
 & 55.0\% & \textgreater{}50 & \textgreater{}50 & 15.8 & 9.3 & 2.6 & 2.8 & 2.4 & \textbf{2.3} \\
\hline
\multirow{2}{*}{jedit} & 52.5\% & 6.5 & \textgreater{}50 & 5.5 & 6.5 & 2.2 & 2.4 & \textbf{2.0} & \textbf{2.0} \\
 & 55.0\% & \textgreater{}50 & \textgreater{}50 & 6.9 & 13.6 & 2.6 & 2.6 & \textbf{2.0 }&\textbf{ 2.0} \\
\hline
\multirow{2}{*}{lucene} & 52.5\% & \textbf{2.0 }& 2.2 & 2.2 & \textgreater{}50 & \textgreater{}50 & \textgreater{}50 &\textbf{ 2.0} & 2.1 \\
 & 55.0\% & \textbf{2.0} & 2.2 & 5.4 & \textgreater{}50 & \textgreater{}50 & \textgreater{}50 & 2.6 & 3.0 \\
\hline
\multirow{2}{*}{xerces} & 35.0\% & \textgreater{}50 & \textgreater{}50 & 3.1 & 9.8 & \textbf{2.0} & 2.2 & 3.2 & 3.2 \\
 & 37.5\% & \textgreater{}50 & \textgreater{}50 & \textgreater{}50 & 14.3 & 2.6 & \textbf{2.4} & 3.6 & 4.0 \\
\hline
\multirow{2}{*}{velocity} & 45.0\% & 2.2 & \textbf{2.0 }& \textgreater{}50 & \textgreater{}50 & \textgreater{}50 &\textgreater{}50 & 2.3 & 2.2 \\
 & 47.5\% &\textbf{ 2.2 }& 3.3 & \textgreater{}50 & \textgreater{}50 & \textgreater{}50 & \textgreater{}50  & 3.2 & 3.7 \\
\hline
\multirow{2}{*}{xalan} & 50.0\% & 2.4 & 2.5 & 15.3 & \textgreater{}50 & 2.2 & 2.4 &\textbf{ 2.0} & 2.1 \\
 & 52.5\% & 2.7 & 2.5 & \textgreater{}50 & \textgreater{}50 & \textbf{2.2} & 2.4 & 6.5 & 8.3 \\
\hline
\multirow{2}{*}{synapse} & 57.5\% & 14.3 & 13.8 & 9.5 & 9.2 & 2.9 & 3.0 & 2.4 & \textbf{2.2} \\
 & 60.0\% & \textgreater{}50 & \textgreater{}50 & \textgreater{}50 & 14.2 & \textgreater{}50 & \textgreater{}50 & 3.0 & \textbf{2.5} \\
\hline
\multirow{2}{*}{log4j} & 62.5\% & \textgreater{}50 & \textgreater{}50 & 13.4 & 9.3 & \textgreater{}50 & \textgreater{}50 & \textbf{2.0} &\textbf{ 2.0} \\
 & 65.0\% & \textgreater{}50 & \textgreater{}50 & 16.7 & 13.0 & \textgreater{}50 & \textgreater{}50 & \textbf{2.7} & 2.8 \\
\hline
\multirow{2}{*}{poi} & 60.0\% & 2.1 & 2.2 & \textgreater{}50 & \textgreater{}50& \textgreater{}50 & \textbf{2.0} & \textbf{2.0} & \textbf{2.0} \\
 & 62.5\% & 2.1 & 2.2 & \textgreater{}50 & \textgreater{}50 & \textgreater{}50 & \textgreater{}50 & \textbf{2.0} & 2.1 \\
\hline
\multirow{2}{*}{ivy} & 35.0\% & \textgreater{}50 & \textgreater{}50 & 7.8 & 6.6 & 2.2 & 2.6 & \textbf{2.0} &\textbf{ 2.0} \\
 & 37.5\% & \textgreater{}50 & \textgreater{}50 & 8.5 & 9.1 & 2.2 & 2.8 & 3.7 & \textbf{3.6 }\\
\hline
\multirow{2}{*}{prop6} & 25.0\% & \textgreater{}50 & \textgreater{}50 & 8.3 & 7.5 & 2.2 & 2.8 & 2.8 & \textbf{2.0} \\
 & 27.5\% & \textgreater{}50 & \textgreater{}50 & 17.5 & 11.5 & \textgreater{}50 & \textbf{2.8} & 4.3 & 3.2 \\
\hline
\multirow{2}{*}{redaktor} & 22.5\% & \textgreater{}50 & \textgreater{}50 & 5.6 & \textbf{2.2 }& \textgreater{}50 & \textgreater{}50 & 3.0 & 3.0 \\
 & 25.0\% & \textgreater{}50 & \textgreater{}50 & 9.4 & \textbf{2.4} & \textgreater{}50 & \textgreater{}50 & 3.0 & 3.0 \\
\hline
\multirow{2}{*}{tomcat} & 30.0\% & \textgreater{}50 & \textgreater{}50 & 7.8 & 7.8 & 2.6 & \textgreater{}50 & \textbf{2.2} & \textbf{2.2} \\ 
 & 32.5\% & \textgreater{}50 & \textgreater{}50 & 8.6 & 15.0 & \textgreater{}50 & \textgreater{}50 & 2.9 & \textbf{2.8} \\\hline
Total &  & 3 & 1 & 0 & 2 & 3 & 3 & 10 & \textbf{13} \\ 
\toprule
\end{tabular}
\end{table*}

\subsection{RQ3: Is our proposed FedDP sensitive to different distillation datasets?}

In this RQ, we select projects with the largest and smallest number of instances to investigate the impact of the selection of distillation datasets on the performance of FedDP. For the distillation datasets, we take the project as the smallest unit, meaning that for projects with multiple versions, their distillation data encompasses data from all versions. Therefore, we chose `redaktor' as the smallest distillation dataset. In our setting, we employ the projects, excluding the distillation project as the training set. The experimental results are shown in Table \ref{tab:table6}. The column `Instances' shows the size of the distillation dataset, and the column `Variation' demonstrates the extent of changes in model performance of FedDP compared to baseline methods. We use bold type to highlight the best results with different FL algorithms.

When selecting the `camel' project, which possesses the largest number of instances, as the distillation dataset, FedDP outperforms the baseline model in terms of Precision, F1, and AUC. Specifically, compared to FLR, FedDP exhibits improvements of 0.05\% in Precision, 5.11\% in Recall, 2.25\% in F1, and 1.30\% in AUC. And FedDP demonstrates enhancements of 0.98\% in Precision, 4.64\% in Recall, 2.34\% in F1, and 1.96\% in AUC compared to Almity. OpenFLR lags behind all methods in all metrics except Recall.

\begin{table*}[!h]
\caption{The performance of FedDP and baseline methods on different distillation datasets.\label{tab:table6}}
\centering
\setlength{\tabcolsep}{1.5pt}
\resizebox{\columnwidth}{!}{
\begin{tabular}{cccccccrrrccccrrr}
\bottomrule
\multirow{2}{*}{ Project  }& 
\multirow{2}{*}{Instances } & 
\multirow{2}{*}{ Metric} & 
\multicolumn{4}{c}{FedAvg} & 
\multicolumn{3}{c}{Variation} & 
\multicolumn{4}{c}{FedProx} & 
\multicolumn{3}{c}{Variation} \\
\cmidrule(r){4-7} \cmidrule(r){8-10} \cmidrule(r){11-14} \cmidrule(r){15-17}  
&  &  & OpenFLR & FLR &Almity& FedDP & vs. OpenFLR & vs. FLR& vs. Almity & OpenFLR & FLR &Almity& FedDP & vs. OpenFLR & vs. FLR& vs. Almity\\ \hline

\multirow{4}{*}{camel} & \multirow{4}{*}{1837} & Precision & 37.55 & 48.96&49.25 & \textbf{49.00} & 11.45 $\uparrow$ & 0.04 $\uparrow$ & 0.25 $\downarrow$& 37.52 & 48.96 & 46.81
&\textbf{49.02} & 11.5 $\uparrow$ & 0.06 $\uparrow$& 2.21 $\uparrow$\\
 & & Recall & \textbf{67.90} & 54.32 &53.25& 59.60 & 8.30 $\downarrow$ & 5.27 $\uparrow$ & 6.35 $\uparrow$& \textbf{68.62} & 54.93 &56.94& 59.87 & 8.75 $\downarrow$ & 4.94 $\uparrow$& 2.93 $\uparrow$\\
 & & F1 & 44.22 & 46.74 &46.55& \textbf{48.93} & 4.71 $\uparrow$ & 2.19 $\uparrow$ & 2.38 $\uparrow$& 44.44 & 46.83 & 46.84&\textbf{49.14} & 4.71 $\uparrow$ & 2.31 $\uparrow$& 2.30 $\uparrow$\\
 & & AUC & 62.00 & 66.71 & 66.21 & \textbf{68.07} & 6.06 $\uparrow$ & 1.36 $\uparrow$ & 1.86 $\uparrow$ & 61.98 & 66.93 &66.12 & \textbf{68.18} & 6.2 $\uparrow$ & 1.25 $\uparrow$& 2.06 $\uparrow$\\
\hline
\multirow{4}{*}{redaktor} & \multirow{4}{*}{176} & Precision & 39.21 & \textbf{48.94} &48.22& 47.07 & 7.86 $\uparrow$ & 1.88 $\downarrow$ & 1.15 $\downarrow$ & 39.24 & \textbf{48.35} &48.49& 47.00 & 7.76 $\uparrow$& 1.35 $\downarrow$ & 1.49 $\downarrow$\\
 & & Recall & 21.74 & 56.66&46.97 & \textbf{61.71} & 39.96 $\uparrow$ & 5.05 $\uparrow$& 4.74 $\uparrow$ & 21.49 & 57.33 &57.62& \textbf{61.63} & 40.14 $\uparrow$& 4.30 $\uparrow$ & 4.01 $\uparrow$\\
 & & F1 & 25.76 & 48.23&48.01 & \textbf{49.06} & 23.30 $\uparrow$ & 0.84 $\uparrow$ & 1.05 $\uparrow$& 25.69 & 48.27 &48.47& \textbf{49.05} & 23.36 $\uparrow$ & 0.78 $\uparrow$& 0.58 $\uparrow$\\
 & & AUC & 54.56 & 67.50 &67.34& \textbf{67.99} & 13.43 $\uparrow$ & 0.49 $\uparrow$& 0.65 $\uparrow$ & 54.53 & 67.48 &67.55& \textbf{67.97} & 13.44 $\uparrow$ & 0.49 $\uparrow$& 0.42 $\uparrow$\\                       
\toprule
\end{tabular}}
\end{table*}

\begin{figure*}[!h]
\centering
\includegraphics[width=\linewidth]{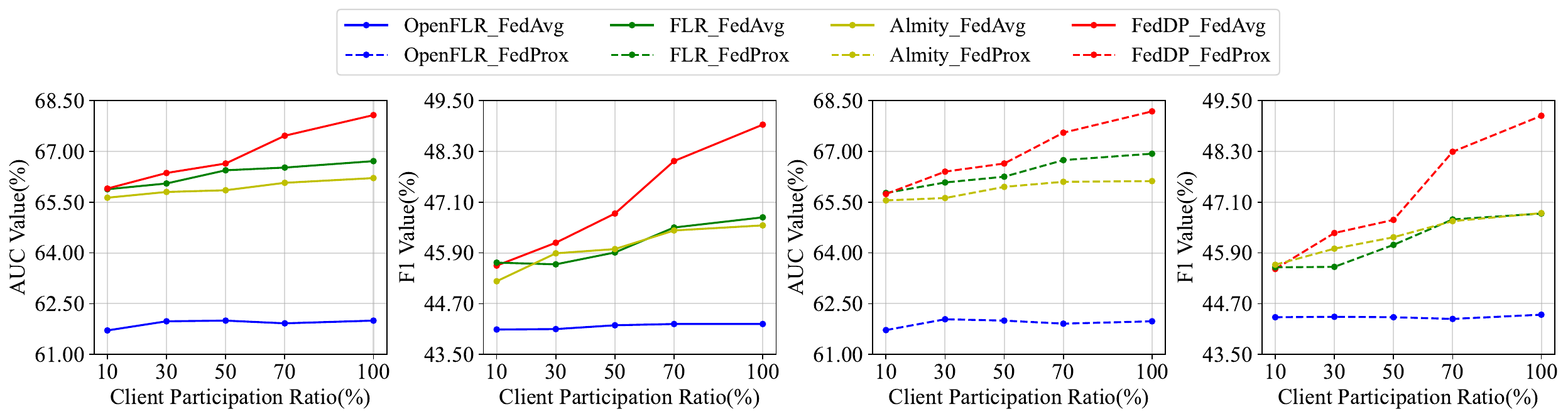}
% \Description{As the number of clients increases, the performance of FedDP and FLR in terms of AUC and F1 scores steadily improves.}
\caption{The performance of FL-based CPDP methods under different Client Participation Ratio $R$.}
\label{fig3}
\end{figure*}

When the `redaktor' project with the fewest instances is selected as the distillation dataset, FedDP maintains its superiority over the baseline model. FedDP outperforms FLR in Recall (by 4.67\%), F1 (by 0.81\%), and AUC (by 0.49\%), albeit with a slight decrease in Precision. When compared to Almity, FedDP exhibits similar magnitudes of performance improvements. It is noteworthy that OpenFLR experiences a severe performance decline in this setting, widening the gap with other methods. This underscores the sensitivity of OpenFLR to the choice of distillation data, further emphasizing the risks associated with directly incorporating open-source data into FL training.

Furthermore, we observe that the distillation with the larger `camel' dataset results in a more significant performance improvement for FedDP over FLR compared to using the `redaktor' dataset. The reason is that larger datasets typically contain more diverse data, representing a wider range of potential data distributions, thus providing richer information and knowledge to enhance the model. FedDP consistently outperforms the baseline models regardless of different distillation projects. This indicates that the knowledge distillation process in FedDP can stably enhance model performance in federated learning, unconstrained by the selection of distillation datasets.

To conclude, FedDP demonstrates robustness in different distillation datasets, maintaining stable performance improvements even in the face of discrepancies among these datasets.

\section{Discussion}\label{dis}
\subsection{Client Participation Ratio}
Client participation ratio $R$ refers to the proportion of clients involved in the training process in each communication round. We explore its impact on the model's predictive performance, as illustrated in Fig. \ref{fig3}. 

We observe that as the client participation ratio increases, the performance of FedDP, Almity, and FLR in terms of AUC and F1 values steadily improves. When the ratio of clients participating in training increases from 10\% to 50\%, the improved performance of FedDP and FLR is relatively increment. However, once more than half of the clients participate in the FL training, there is a marked improvement in the performance of both FedDP and FLR. We attribute this improvement primarily to the richer knowledge from multiple clients and facilitate better model training. Conversely, the performance of OpenFLR almost remains unchanged regardless of the increase in client participation ratio due to the direct training on open-source project data, which harms model performance and prevents it from optimizing from the optimal direction.

\begin{figure*}[!h]
\centering
% \Description{FedDP is not sensitive to the distillation steps and Sampling Sizes.}
\includegraphics[width=\linewidth]{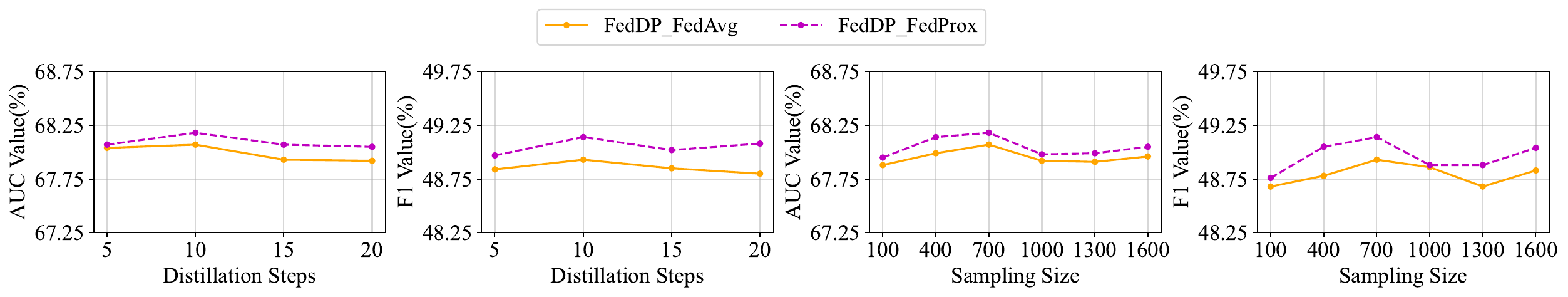}
\caption{The performance of FedDP on different Distillation Steps $N$ and Sampling Sizes $p$.}
\label{fig4}
\end{figure*}

\begin{table*}[!h]
\caption{The ablation study of two solutions in FedDP.\label{tab:table7}}
\centering
\begin{tabular}{clcccc}
\bottomrule
Algorithms & \multicolumn{1}{c}{Methods}   & Precision & Recall & F1 & AUC\\ \hline
\multirow{3}{*}{FedAvg} & FedDP   & \textbf{49.00} & \textbf{59.60} &  \textbf{48.93} & \textbf{68.07} \\
 & (w/o) factor   & 48.91 & 59.11 &48.63 & 67.80\\
 & (w/o) factor \& distill   &48.96 & 54.32 & 46.74 & 66.71 \\ \hline
\multirow{3}{*}{FedProx} & FedDP   & \textbf{49.02} & \textbf{59.87} & \textbf{49.14} & \textbf{68.18} \\
 & (w/o) factor  & 48.80 & 59.50 & 48.60 & 67.87 \\
 & (w/o) factor \& distill   &48.96 & 54.93 & 46.83 & 66.93 \\ \toprule
\end{tabular}
\end{table*}

\subsection{Distillation Step and Sampling Size}

The knowledge distillation process is the key technique of FedDP. During each round of distillation, we randomly sample a portion of the data from the distillation dataset according to the sampling size $p$. The number of distillation steps $N$ refers to the distillation epochs at the server. We explore the impact of these two parameters on the performance of FedDP, as depicted in Fig. \ref{fig4}.

As the number of distillation steps varies, the performance of FedDP remains generally stable. When the distillation step is set to 10, the values of F1 and AUC reach their maximum. With FedAvg, AUC and F1 values vary by a max of 0.15\% and 0.13\%. The maximum differences with FedProx in AUC and F1 are 0.13\% and 0.17\% respectively. Therefore, we suggest that FedDP is not sensitive to the distillation steps.

The project `camel' comprises 1837 instances. Thus, our experiment varies the sampling size with values from the set [100, 400, 700, 1000, 1300, 1600]. As the sampling size $p$ changed from 100 to 1600, both the AUC and F1 values of FedDP show minor fluctuations, peaking at $p = 700$. The max differences in AUC and F1 are 0.18\% and 0.25\% with FedAvg while 0.23\% and 0.39\% with FedProx. Taking into account both performance and training overhead, we set $p = 700$ in our experiments.

\subsection{Ablation Study}
In view of that FedDP has two components: Local Heterogeneity Awareness and Global Knowledge Distillation. It is necessary to validate the effectiveness of these two components. Here we conduct ablation experiments on FedDP, with results shown in Table \ref{tab:table7}. Here, we abbreviate the correlation factor as `factor' while `distill' represents the federated knowledge distillation technique.

Without the support of Local Heterogeneity Awareness and Global Knowledge Distillation, FedDP would be reduced to the FLR method. We can observe that the knowledge distillation technique plays an important role in enhancing model performance, demonstrating improvements of 0.05\%, 5.11\%, 2.25\%, and 1.30\% in Precision, Recall, F1, and AUC. This underscores the capability of our proposed solutions to increase Recall performance while maintaining Precision. A high Recall value is crucial for CPDP as it ensures the identification of more potential software defects, which could otherwise lead to software failures or security vulnerabilities.

Comparing FedDP with the `(w/o) factor', the performance of the model gets an increment improvement when the correlation factors are introduced in the distillation process. This suggests that our Local Heterogeneity Awareness approach can identify the correlation between distillation data and local data, enabling the server to generate a superior teacher model and optimize Global Knowledge Distillation.

To sum up, the Global Knowledge Distillation approach contributes greatly to the performance enhancement of FedDP, and the utilization of the correlation factors can further optimize the distillation process and thus enhance the predictive performance.

\subsection{Generalization}
The FedDP framework excels in CPDP, and its exceptional generalization capability allows it to be applied to other tasks requiring privacy protection, such as code smell detection, code clone detection, and classification of bug-fix commits. Taking code clone detection as an example, this task aims to find similar or functionally equivalent code snippets, which are crucial for improving software quality and reducing maintenance. Codebases from different organizations or projects often exhibit unique styles, leading to data distribution diversity. With FedDP, organizations can collaboratively train efficient models without exposing their codebases.

FedDP can effectively address the Non-IID issue, fostering knowledge sharing and collaboration across organizations while strictly protecting data privacy. This opens up a new path for research and practice in the field of software engineering.

\section{Threats to Validity}
\label{threat}
\subsection{Internal Validity }
Not all baseline methods have open-sourced their code. We strictly adhere to the descriptions in the relevant papers to replicate the method, but it is still possible that we may not be able to fully implement all the details of the original works, which may affect our assessment of the model's effectiveness.
\subsection{External Validity}
In this study, we extensively research 14 Java projects in the Promise repository and 5 C projects in the Softlab repository. These projects cover a sufficiently diverse range of data to construct and evaluate FedDP. However, there may be some uncertainty regarding the performance of FedDP when applied to projects utilizing other programming languages. In the future, we will try to use more datasets to minimize this threat. 

\subsection{Conclusion Validity}
The validity of conclusions mainly involves whether the evaluation metrics used are reasonable. Class imbalance is a prevalent challenge in defect prediction tasks. The F1 value can balance the recall and precision of defect class predictive performance, while the AUC metric can comprehensively evaluate the model's classification performance. Both of these are crucial evaluation indicators commonly used in the defect prediction domain.

\section{Conclusion and Future Work}\label{concl}
In this paper, we propose a realistic and challenging problem, privacy-preserving CPDP with data heterogeneity. To tackle this issue, we first intuitively try to consolidate the model with additional training on the open-source project data. Then, we empirically analyze the failure and improve with FedDP. FedDP employs the open-source project data as the distillation data and enhances the global model via heterogeneity-aware model ensemble knowledge distillation. Extensive experiments conducted on various settings and baselines show that FedDP achieves significant improvement in terms of AUC and F1 values respectively. 

Although our method has demonstrated great effectiveness over the privacy-preserving CPDP, the extent to which FedDP improves model performance still depends on the quality of the distillation dataset. To deploy the privacy-preserving CPDP in practical settings, it is necessary to consider data-free knowledge distillation techniques such as the generative model. In the future, we seek to work a step forward in this field.
\bmsection*{DATA AVAILABILITY STATEMENT}
The data that support the findings of this study are openly available in FedDP(\href{https://github.com/waiwaiwang/FedDP.git}{https://github.com/waiwaiwang/FedDP.git}).
\bmsection*{Conflict of interest}
The authors declare no potential conflict of interests.
\bmsection*{ORCID}
\textit{Yichen Li} \orcidlink{0009-0009-8630-2504}
\href{https://orcid.org/0009-0009-8630-2504}{https://orcid.org/0009-0009-8630-2504}\\
\textit{Xiaofang Zhang} \orcidlink{0000-0002-8667-0456}
\href{https://orcid.org/0000-0002-8667-0456}{https://orcid.org/0000-0002-8667-0456}

\bibliography{ref}

\begin{thebibliography}{10}
\providecommand \doibase [0]{http://dx.doi.org/}%

\bibitem{SDP1}
Ostrand TJ, Weyuker EJ, Bell RM. Predicting the Location and Number of Faults in Large Software Systems. {\it IEEE Transactions on Software Engineering.} 2005\string;31(4)\string:340--355.
\newblock \href {\doibase 10.1109/TSE.2005.49} {doi: 10.1109/TSE.2005.49}

\bibitem{SDP2}
Lessmann S, Baesens B, Mues C, Pietsch S. Benchmarking Classification Models for Software Defect Prediction: {A} Proposed Framework and Novel Findings. {\it IEEE Transactions on Software Engineering.} 2008\string;34(4)\string:485--496.
\newblock \href {\doibase 10.1109/TSE.2008.35} {doi: 10.1109/TSE.2008.35}

\bibitem{SDP3}
Hall T, Beecham S, Bowes D, Gray D, Counsell S. A Systematic Literature Review on Fault Prediction Performance in Software Engineering. {\it IEEE Transactions on Software Engineering.} 2012\string;38(6)\string:1276-1304.
\newblock \href {\doibase 10.1109/TSE.2011.103} {doi: 10.1109/TSE.2011.103}

\bibitem{WPDP1}
Menzies T, Greenwald J, Frank A. Data Mining Static Code Attributes to Learn Defect Predictors. {\it IEEE Transactions on Software Engineering.} 2007\string;33(1)\string:2-13.
\newblock \href {\doibase 10.1109/TSE.2007.256941} {doi: 10.1109/TSE.2007.256941}

\bibitem{WPDP2}
Dalla~Palma S, Di~Nucci D, Palomba F, Tamburri DA. Within-Project Defect Prediction of Infrastructure-as-Code Using Product and Process Metrics. {\it IEEE Transactions on Software Engineering.} 2022\string;48(6)\string:2086-2104.
\newblock \href {\doibase 10.1109/TSE.2021.3051492} {doi: 10.1109/TSE.2021.3051492}

\bibitem{WPDP3}
Jing XY, Wu F, Dong X, Xu B. An Improved SDA Based Defect Prediction Framework for Both Within-Project and Cross-Project Class-Imbalance Problems. {\it IEEE Transactions on Software Engineering.} 2017\string;43(4)\string:321-339.
\newblock \href {\doibase 10.1109/TSE.2016.2597849} {doi: 10.1109/TSE.2016.2597849}

\bibitem{cpdp_nec1}
He Z, Shu F, Yang Y, Li M, Wang Q. An investigation on the feasibility of cross-project defect prediction. {\it Automated Software Engineering.} 2012\string;19(2)\string:167-199.
\newblock \href {\doibase 10.1007/s10515-011-0090-3} {doi: 10.1007/s10515-011-0090-3}

\bibitem{CPDP1}
Turhan B, Menzies T, Bener AB, Stefano JD. On the relative value of cross-company and within-company data for defect prediction. {\it Empirical Software Engineering.} 2009\string;14(5)\string:540-578.
\newblock \href {\doibase 10.1007/S10664-008-9103-7} {doi: 10.1007/S10664-008-9103-7}

\bibitem{cpdp_nec3}
Kitchenham BA, Mendes E, Travassos GH. Cross versus Within-Company Cost Estimation Studies: {A} Systematic Review. {\it IEEE Transactions on Software Engineering.} 2007\string;33(5)\string:316--329.
\newblock \href {\doibase 10.1109/TSE.2007.1001} {doi: 10.1109/TSE.2007.1001}

\bibitem{CPDP2}
Xia X, Lo D, Pan SJ, Nagappan N, Wang X. HYDRA: Massively Compositional Model for Cross-Project Defect Prediction. {\it IEEE Transactions on Software Engineering.} 2016\string;42(10)\string:977-998.
\newblock \href {\doibase 10.1109/TSE.2016.2543218} {doi: 10.1109/TSE.2016.2543218}

\bibitem{CPDP3}
Canfora G, Lucia AD, Penta MD, Oliveto R, Panichella A, Panichella S. Multi-objective Cross-Project Defect Prediction. In: {IEEE} Computer Society. 2013\string:252--261

\bibitem{cpdp_nec2}
Hosseini S, Turhan B, Gunarathna D. A Systematic Literature Review and Meta-Analysis on Cross Project Defect Prediction. {\it IEEE Transactions on Software Engineering.} 2019\string;45(2)\string:111--147.
\newblock \href {\doibase 10.1109/TSE.2017.2770124} {doi: 10.1109/TSE.2017.2770124}

\bibitem{Weyuker}
Weyuker EJ, Ostrand TJ, Bell RM. Do too many cooks spoil the broth? Using the number of developers to enhance defect prediction models. {\it Empirical Software Engineering.} 2008\string;13(5)\string:539-559.
\newblock \href {\doibase 10.1007/s10664-008-9082-8} {doi: 10.1007/s10664-008-9082-8}

\bibitem{Li_loc}
Li Z, Jing XY, Zhu X, Zhang H, Xu B, Ying S. On the Multiple Sources and Privacy Preservation Issues for Heterogeneous Defect Prediction. {\it IEEE Transactions on Software Engineering.} 2019\string;45(4)\string:391-411.
\newblock \href {\doibase 10.1109/TSE.2017.2780222} {doi: 10.1109/TSE.2017.2780222}

\bibitem{PetersLACE2}
Peters F, Menzies T, Layman L. {LACE2:} Better Privacy-Preserving Data Sharing for Cross Project Defect Prediction. In: Proceedings of the 37th International Conference on Software Engineering, {ICSE} 2015, Florence, Italy. 2015\string:801--811

\bibitem{Yamamoto}
Yamamoto H, Wang D, Rajbahadur GK, Kondo M, Kamei Y, Ubayashi N. Towards Privacy Preserving Cross Project Defect Prediction with Federated Learning. In: {IEEE} International Conference on Software Analysis, Evolution and Reengineering, {SANER} 2023, Taipa, Macao. 2023\string:485--496

\bibitem{xiaxin}
Yang Y, Hu X, Gao Z, et al. Federated Learning for Software Engineering: A Case Study of Code Clone Detection and Defect Prediction. {\it IEEE Transactions on Software Engineering.} 2024\string;50(2)\string:296-321.
\newblock \href {\doibase 10.1109/TSE.2023.3347898} {doi: 10.1109/TSE.2023.3347898}

\bibitem{Shriram}
Shanbhag S, Chimalakonda S. Exploring the under-explored terrain of non-open source data for software engineering through the lens of federated learning. In: Proceedings of the 30th {ACM} Joint European Software Engineering Conference and Symposium on the Foundations of Software Engineering, {ESEC/FSE} 2022, Singapore. 2022\string:1610--1614

\bibitem{Sai}
Karimireddy SP, Kale S, Mohri M, Reddi SJ, Stich SU, Suresh AT. {SCAFFOLD:} Stochastic Controlled Averaging for Federated Learning. In: Proceedings of the 37th International Conference on Machine Learning, {ICML} 2020, Virtual Event. 2020\string:5132--5143.

\bibitem{Xiang}
Li X, Huang K, Yang W, Wang S, Zhang Z. On the Convergence of FedAvg on Non-IID Data. In: 8th International Conference on Learning Representations, {ICLR} 2020, Addis Ababa, Ethiopia. 2020.

\bibitem{He}
He Z, Peters F, Menzies T, Yang Y. Learning from Open-Source Projects: An Empirical Study on Defect Prediction. In: 2013 ACM/IEEE International Symposium on Empirical Software Engineering and Measurement. 2013\string:45-54

\bibitem{back_cpdp1}
Ni C, Xia X, Lo D, Chen X, Gu Q. Revisiting Supervised and Unsupervised Methods for Effort-Aware Cross-Project Defect Prediction. {\it IEEE Transactions on Software Engineering.} 2022\string;48(3)\string:786--802.
\newblock \href {\doibase 10.1109/TSE.2020.3001739} {doi: 10.1109/TSE.2020.3001739}

\bibitem{back_cpdp2}
Zhang F, Zheng Q, Zou Y, Hassan AE. Cross-project defect prediction using a connectivity-based unsupervised classifier. In: Proceedings of the 38th International Conference on Software Engineering, {ICSE} 2016, Austin, TX, USA. 2016\string:309--320

\bibitem{back_cpdp3}
Jin C. Cross-project software defect prediction based on domain adaptation learning and optimization. {\it Expert Systems with Applications.} 2021\string;171\string:114637.
\newblock \href {\doibase 10.1016/J.ESWA.2021.114637} {doi: 10.1016/J.ESWA.2021.114637}

\bibitem{master}
Tong H, Zhang D, Liu J, et al. {MASTER:} Multi-Source Transfer Weighted Ensemble Learning for Multiple Sources Cross-Project Defect Prediction. {\it IEEE Transactions on Software Engineering.} 2024\string;50(5)\string:1281--1305.
\newblock \href {\doibase 10.1109/TSE.2024.3381235} {doi: 10.1109/TSE.2024.3381235}

\bibitem{tca+}
Nam J, Pan SJ, Kim S. Transfer defect learning. In: Proceedings of the 35th International Conference on Software Engineering, {ICSE} 2013,San Francisco, CA, USA. 2013\string:382--391

\bibitem{tnb}
Ma Y, Luo G, Zeng X, Chen A. Transfer learning for cross-company software defect prediction. {\it Information and Software Technology.} 2012\string;54(3)\string:248--256.
\newblock \href {\doibase 10.1016/J.INFSOF.2011.09.007} {doi: 10.1016/J.INFSOF.2011.09.007}

\bibitem{hydra}
Xia X, Lo D, Pan SJ, Nagappan N, Wang X. {HYDRA:} Massively Compositional Model for Cross-Project Defect Prediction. {\it IEEE Transactions on Software Engineering.} 2016\string;42(10)\string:977--998.
\newblock \href {\doibase 10.1109/TSE.2016.2543218} {doi: 10.1109/TSE.2016.2543218}

\bibitem{wang2023fedcda}
Wang H, Xu H, Li Y, Xu Y, Li R, Zhang T. FedCDA: Federated Learning with Cross-rounds Divergence-aware Aggregation. In: The Twelfth International Conference on Learning Representations.  2023.

\bibitem{li2024unleashing}
Li Y, Wang H, Xu W, et al. Unleashing the Power of Continual Learning on Non-Centralized Devices: A Survey. {\it IEEE Communications Surveys \& Tutotials.} 2024.

\bibitem{zhu}
Zhu H, Zhang H, Jin Y. From federated learning to federated neural architecture search: a survey. {\it Complex \& Intelligent Systems.} 2021\string;7\string:639--657.

\bibitem{yang}
Yang Q, Liu Y, Chen T, Tong Y. Federated machine learning: Concept and applications. {\it ACM Transactions on Intelligent Systems and Technology.} 2019\string;10(2)\string:1--19.

\bibitem{Li_2024_CVPR}
Li Y, Li Q, Wang H, Li R, Zhong W, Zhang G. Towards Efficient Replay in Federated Incremental Learning. In: Proceedings of the IEEE/CVF Conference on Computer Vision and Pattern Recognition (CVPR).  2024\string:12820-12829.

\bibitem{li_tpds}
Li Y, Xu W, Qi Y, Wang H, Li R, Guo S. SR-FDIL: Synergistic Replay for Federated Domain-Incremental Learning. {\it IEEE Transactions on Parallel and Distributed Systems.} 2024\string;35(11)\string:1879-1890.
\newblock \href {\doibase 10.1109/TPDS.2024.3436874} {doi: 10.1109/TPDS.2024.3436874}

\bibitem{fedavg}
McMahan B, Moore E, Ramage D, Hampson S, Arcas yBA. Communication-Efficient Learning of Deep Networks from Decentralized Data. In: Proceedings of the 20th International Conference on Artificial Intelligence and Statistics, {AISTATS} 2017, Fort Lauderdale, {USA}. 2017.

\bibitem{back_fl1}
Jeong E, Oh S, Kim H, Park J, Bennis M, Kim S. Communication-Efficient On-Device Machine Learning: Federated Distillation and Augmentation under Non-IID Private Data. {\it CoRR.} 2018\string;abs/1811.11479.

\bibitem{backfl2}
Liu L, Zhang J, Song S, Letaief KB. Client-Edge-Cloud Hierarchical Federated Learning. In: {IEEE} International Conference on Communications, {ICC} 2020, Dublin, Ireland. 2020\string:1--6

\bibitem{li2024rehearsal}
Li Y, Wang Y, Xiao T, Wang H, Qi Y, Li R. Rehearsal-Free Continual Federated Learning with Synergistic Regularization. In: The Thirteenth International Conference on Learning Representations.  2024.

\bibitem{FedProx}
Li T, Sahu AK, Zaheer M, Sanjabi M, Talwalkar A, Smith V. Federated Optimization in Heterogeneous Networks. In: Proceedings of the Third Conference on Machine Learning and Systems, MLSys 2020, Austin, TX, USA. 2020.

\bibitem{KDLOGIT1}
Hinton GE, Vinyals O, Dean J. Distilling the Knowledge in a Neural Network. {\it CoRR.} 2015\string;abs/1503.02531.

\bibitem{KD2}
Li Y, Yang J, Song Y, Cao L, Luo J, Li L. Learning from Noisy Labels with Distillation. In: {IEEE} International Conference on Computer Vision, {ICCV} 2017, Venice, Italy. 2017\string:1928--1936

\bibitem{KD1}
Phuong M, Lampert C. Distillation-Based Training for Multi-Exit Architectures. In: {IEEE/CVF} International Conference on Computer Vision, {ICCV} 2019, Seoul, Korea (South). 2019\string:1355--1364

\bibitem{KD3}
Yang C, Xie L, Qiao S, Yuille AL. Training Deep Neural Networks in Generations: {A} More Tolerant Teacher Educates Better Students. In: The Thirty-Third {AAAI} Conference on Artificial Intelligence, {AAAI} 2019, Honolulu, Hawaii, USA. 2019\string:5628--5635

\bibitem{fkd1}
Guo Q, Wang X, Wu Y, et al. Online Knowledge Distillation via Collaborative Learning. In: {IEEE/CVF} Conference on Computer Vision and Pattern Recognition, {CVPR} 2020, Seattle, WA, USA. 2020\string:11017--11026

\bibitem{fkd2}
Wu G, Gong S. Peer Collaborative Learning for Online Knowledge Distillation. In: The Thirty-Fifth {AAAI} Conference on Artificial Intelligence, {AAAI} 2021, Virtual Event. 2021\string:10302--10310

\bibitem{FKDDF}
Lin T, Kong L, Stich SU, Jaggi M. Ensemble distillation for robust model fusion in federated learning. In: Proceedings of the 34th International Conference on Neural Information Processing Systems,NeurIPS 2020, virtual Event. 2020.

\bibitem{fedgen}
Zhu Z, Hong J, Zhou J. Data-Free Knowledge Distillation for Heterogeneous Federated Learning. In: Proceedings of the 38th International Conference on Machine Learning, {ICML} 2021,Virtual Event. 2021\string:12878--12889.

\bibitem{dafkd}
Wang H, Li Y, Xu W, Li R, Zhan Y, Zeng Z. DaFKD: Domain-aware Federated Knowledge Distillation. In: {IEEE/CVF} Conference on Computer Vision and Pattern Recognition, {CVPR} 2023, Vancouver, BC, Canada. 2023\string:20412--20421

\bibitem{method_kd2}
Xie Q, Luong M, Hovy EH, Le QV. Self-Training With Noisy Student Improves ImageNet Classification. In: {IEEE/CVF} Conference on Computer Vision and Pattern Recognition, {CVPR} 2020, Seattle, WA, USA. 2020\string:10684--10695

\bibitem{Kl}
Kullback S, Leibler RA. On Information and Sufficiency. {\it Annals of Mathematical Statistics.} 1951\string;22\string:79-86.

\bibitem{promise1}
Wang S, Liu T, Nam J, Tan L. Deep Semantic Feature Learning for Software Defect Prediction. {\it IEEE Transactions on Software Engineering.} 2020\string;46(12)\string:1267--1293.
\newblock \href {\doibase 10.1109/TSE.2018.2877612} {doi: 10.1109/TSE.2018.2877612}

\bibitem{promise2}
Okutan A, Yildiz OT. Software defect prediction using Bayesian networks. {\it Empirical Software Engineering.} 2014\string;19(1)\string:154--181.
\newblock \href {\doibase 10.1007/S10664-012-9218-8} {doi: 10.1007/S10664-012-9218-8}

\bibitem{metric1}
Wang S, Liu T, Tan L. Automatically learning semantic features for defect prediction. In: Proceedings of the 38th International Conference on Software Engineering, {ICSE} 2016, Austin, TX, USA. 2016\string:297--308

\bibitem{metric2}
Zhang F, Mockus A, Keivanloo I, Zou Y. Towards building a universal defect prediction model with rank transformed predictors. {\it Empirical Software Engineering.} 2016\string;21\string:2107--2145.

\bibitem{metric3}
Zhang F, Keivanloo I, Zou Y. Data transformation in cross-project defect prediction. {\it Empirical Software Engineering.} 2017\string;22\string:3186--3218.

\bibitem{xztest1}
Zhuang W, Wang H, Zhang X. Just-in-time defect prediction based on {AST} change embedding. {\it Knowledge-Based Systems.} 2022\string;248\string:108852.
\newblock \href {\doibase 10.1016/J.KNOSYS.2022.108852} {doi: 10.1016/J.KNOSYS.2022.108852}

\end{thebibliography}

\end{document}